\documentclass[fleqn,10pt]{wlscirep}
\usepackage[utf8]{inputenc}
\usepackage[T1]{fontenc}

\usepackage[numbers]{natbib}

\usepackage{xr-hyper}
\makeatletter
\newcommand*{\addFileDependency}[1]{
  \typeout{(#1)}
  \@addtofilelist{#1}
  \IfFileExists{#1}{}{\typeout{No file #1.}}
}
\makeatother

\newcommand*{\myexternaldocument}[1]{%
    \externaldocument{#1}%
    \addFileDependency{#1.tex}%
    \addFileDependency{#1.aux}%
}
\myexternaldocument{suppl}

\usepackage{color,soul}
\usepackage{algorithm,algpseudocode}
\newcommand{\angstrom}{\textup{\AA}}

\usepackage{setspace} 
\usepackage{booktabs}
\usepackage{multirow}
\usepackage[labelfont=bf,labelsep=period,justification=raggedright]{caption}

\usepackage[capitalize,noabbrev]{cleveref}

\title{Multi-Objective Latent Space Optimization of Generative Molecular Design Models}

\author[1]{A N M Nafiz Abeer}
\author[2]{Nathan M. Urban}
\author[3]{M Ryan Weil}
\author[4]{Francis J Alexander}
\author[1,2,*]{Byung-Jun Yoon}

\affil[1]{Department of Electrical and Computer Engineering, Texas A\&M University, College Station, TX 77843, USA}
\affil[2]{Computational Science Initiative, Brookhaven National Laboratory, Upton, NY 11973, USA}
\affil[3]{Strategic and Data Science Initiatives, Frederick National Laboratory, Frederick, MD 21702, USA}
\affil[4]{Computing, Environment and Life Sciences, Argonne National Laboratory, Lemont, IL 60439, USA}
\affil[*]{Correspondence: bjyoon@ece.tamu.edu}

\begin{abstract}
Molecular design based on generative models, such as variational autoencoders (VAEs), has become increasingly popular in recent years due to its efficiency for exploring high-dimensional molecular space to identify molecules with desired properties. While the efficacy of the initial model strongly depends on the training data, the sampling efficiency of the model for suggesting novel molecules with enhanced properties can be further enhanced via latent space optimization. In this paper, we propose a multi-objective latent space optimization (LSO) method that can significantly enhance the performance of generative molecular design (GMD). The proposed method adopts an iterative weighted retraining approach, where the respective weights of the molecules in the training data are determined by their Pareto efficiency. We demonstrate that our multi-objective GMD LSO method can significantly improve the performance of GMD for jointly optimizing multiple molecular properties.
\end{abstract}
\begin{document}

\flushbottom
\maketitle
%
%
\thispagestyle{empty}
\section*{Introduction}


The development of quantitative structure–activity relationship (QSAR)\cite{QSAR} models has accelerated the drug design process. However, designing molecules with the desired drug properties through direct optimization over the chemical space remains challenging due to the high dimensionality of the domain. While drug discovery based on high-throughput screening (HTS) systems~\cite{HTS} has been shown to be highly useful, the computational cost needed for screening a huge candidate pool is formidable. Additionally, the design and operation of computational HTS pipelines have traditionally relied on expert intuition and various heuristics, resulting in sub-optimal performance~\cite{Woo2021OCC, Woo2022OCC-RP}. Furthermore, should one wish to consider drug candidates beyond the pool of known drugs and drug-like molecules, expanding the pool faces the same challenges of molecular design in high-dimensional chemical space. As drug discovery involves consideration and optimization of multiple properties, which may conflict one another, this multi-objective optimization aspect further exacerbates the aforementioned design challenges. 

Recent advances in deep generative models provide promising alternatives to conventional computational approaches for drug discovery, which may be able to effectively address many of these challenges.
A representative example is the work by G\'{o}mez-Bombarelli et al.~\cite{Bombarelli_cont}, in which they prosed the use of a variational auto-encoder (VAE) to convert the input molecules, originally represented by SMILES strings, into a continuous lower-dimensional representation in a latent space.
This approach effectively maps molecules in the original chemical space, which is high-dimensional and discrete, to a latent space, which is low-dimensional and continuous, thereby enabling efficient numerical optimization in the latent space in pursuit of molecules with enhanced target attributes. In this study~\cite{Bombarelli_cont}, a Gaussian Process (GP) was used to model and optimize the property predictor in the latent space, which was shown to significantly outperform molecular optimization in the original chemical space using genetic algorithm (GA) as well as randomized search in the latent space. Winter et al.~\cite{PSO_cont_latent_space} adopted Particle Swarm Optimization (PSO), instead of   Bayesian optimization (BO), aiming at further improving the computational efficiency for multi-objective molecular optimization, also in the latent space. In this work, multiple properties of interest were jointly optimized by defining a single objective function through scalarization via weighted combination of multiple property scores. 
As noted in~\cite{Bombarelli_cont}, the generative model may not always suggest molecules with valid molecular structures, which may degrade the overall efficiency of the generative molecular design (GMD) approach. Empirically, this phenomenon has been shown to occur when data points representing the molecules are sampled in regions of the latent space that are far away from the region where the original training data were located. 
To deal with this shortcoming, the search for an optimized molecule with desirable attributes can be formulated as a constrained BO problem~\cite{constrained_BO}, which has been shown to improve the validity of the novel molecules produced by the generative model. The junction-tree VAE (JT-VAE)~\cite{JTVAE-paper} tackles this issue by taking a two-phase approach. In the first phase, the JT-VAE generates a junction-tree that represents the overall scaffold for a molecular graph, which specifies the relative arrangement of valid subgraph structures learned from the training data. During the second phase, subgraphs corresponding to chemical substructures are combined according to the junction-tree to obtain the final molecular graph. As a result, JT-VAE is capable of suggesting novel molecules in the latent space that can be decoded into legitimate molecules with a high chance. In order to explore the latent space to produce novel molecules with targeted attributes, the VAE may also be conditioned by the desired property values. For example, Kang et al.~\cite{conditional_gen} 
proposed a semi-supervised VAE (SSVAE), which simultaneously performs  property prediction and molecular generation, resulting in conditioning the model such that it suggests molecules in the latent space that are centered around a desired range of properties.


In addition to the aforementioned schemes that perform molecular optimization in the latent space, another popular strategy is to first train a generative network to model the input data distribution, which is followed by fine-tuning the model via reinforcement learning (RL) to meet the design criteria. The work by Olivecrona et al.~\cite{olivecrona2017} proposed a fine-tuning approach that facilitates the generation of high-scoring molecules without deviating away from the original input data distribution. Shi et al.~\cite{shi2020graphaf} applied an RL-based policy to fine-tune a generative network model for molecular graphs. In the objective-reinforced generative adversarial network (ORGAN)~\cite{ORGANIC} framework, the reward for generating molecules with better properties is assimilated into the loss function to guide the latent space distribution during model training. To reduce the potential bias in the generative network that may arise from the training dataset, Zhou et al.~\cite{MolDQN} proposed Molecule Deep Q-Networks (MolDQN). In this work, the molecular generation problem was formulated as a Markov Decision Process (MDP) and a Deep Q-network (DQN)~\cite{deepQ_network} was used to find the optimal design policy for the given MDP. The allowable actions in the MDP are dictated by relevant domain knowledge (i.e., chemical reactions) to ensure the validity of the generated molecule. In order to jointly optimize multiple molecular attributes, MolDQN also resorted to scalarization by defining a single objective function based on a weighted combination of multiple property scores.

Like other data-driven models, the initial capability of the generative molecular models to suggest novel molecules will be determined--at least to a certain extent--by the training data. In fact, the generated molecules are likely to reside in a similar chemical space as the molecules in the original training set, which may make it challenging to design new candidate molecules whose target attributes significantly exceed those of the original molecules--regardless of whether we perform molecular optimization in the latent space~\cite{Bombarelli_cont, JTVAE-paper, conditional_gen} or adopt a fine-tuning strategy~\cite{olivecrona2017, shi2020graphaf, ORGANIC, MolDQN}. 
To mitigate this issue, a number of recent efforts aimed to improve the generative models by incorporating additional training data, generated either from experiments or simulations, where the goal is to ensure that these models are primed to suggest novel molecules with enhanced target properties that go beyond the initial molecules.
Yang et al.~\cite{Stochastic_iterative} proposed an iterative retraining approach to improve the quality of the molecules sampled from the latent space of the generative model. In this approach, they pre-train the generative model jointly with a property predictor. The trained model is used to generate novel molecules from which a small batch of molecules with the best properties are selected using the predictor and added to the training dataset. The extended dataset is subsequently used for retraining to update the latent space to make it more amenable to produce better molecules with improved properties.

A similar effort from Iovanac et al.~\cite{iovanac2022actively} utilizes the grammar variational autoencoder~\cite{kusner2017grammar}, jointly trained with a linear predictor network to sample new molecules along the latent dimension corresponding to the targeted property region. The new molecules go through further screening to be used with the training dataset to retrain the generative model iteratively. 
In Liu et al.~\cite{chance-constrained}, the generator in a generative adversarial network (GAN) is iteratively updated under a chance-constrained optimization framework. They employ a validity function to guide the optimization of the property value within the region of grammatically correct input sequences or legitimate structures. A recent work by Tripp et al.~\cite{tripp2020sampleefficient} proposed a weighted retraining approach to reshape the latent space of a VAE to make it more sampling-efficient for producing novel molecules with improved properties. For this purpose, the weights assigned to the data points in the training dataset are determined by the rank of the corresponding molecules according to the objective function that evaluates the property of interest.  
%
%
Another recent iterative approach~\cite{active_transfer_learning} applies a genetic algorithm (GA) along with domain knowledge to generate a new set of potentially improved candidates from the initial training data. The property values of these candidates are validated by high-throughput experiments or simulations. A deep neural network (DNN) property predictor is then retrained using a larger dataset augmented by the candidates generated by a GA. In contrast to the generative approaches~\cite{Stochastic_iterative, chance-constrained, tripp2020sampleefficient,iovanac2022actively}, the candidates produced by a GA generally remain closer to the training data. As a result, the predictions from the DNN remain relatively reliable but the novelty of the produced candidates tends to be limited. Additionally, the (computational) cost of the experiments or  simulations to assess the property of the novel candidates can be a significant burden when the candidate pool gets larger.

\begin{figure}[!ht]
    \centering
    \includegraphics[width=0.75\textwidth]{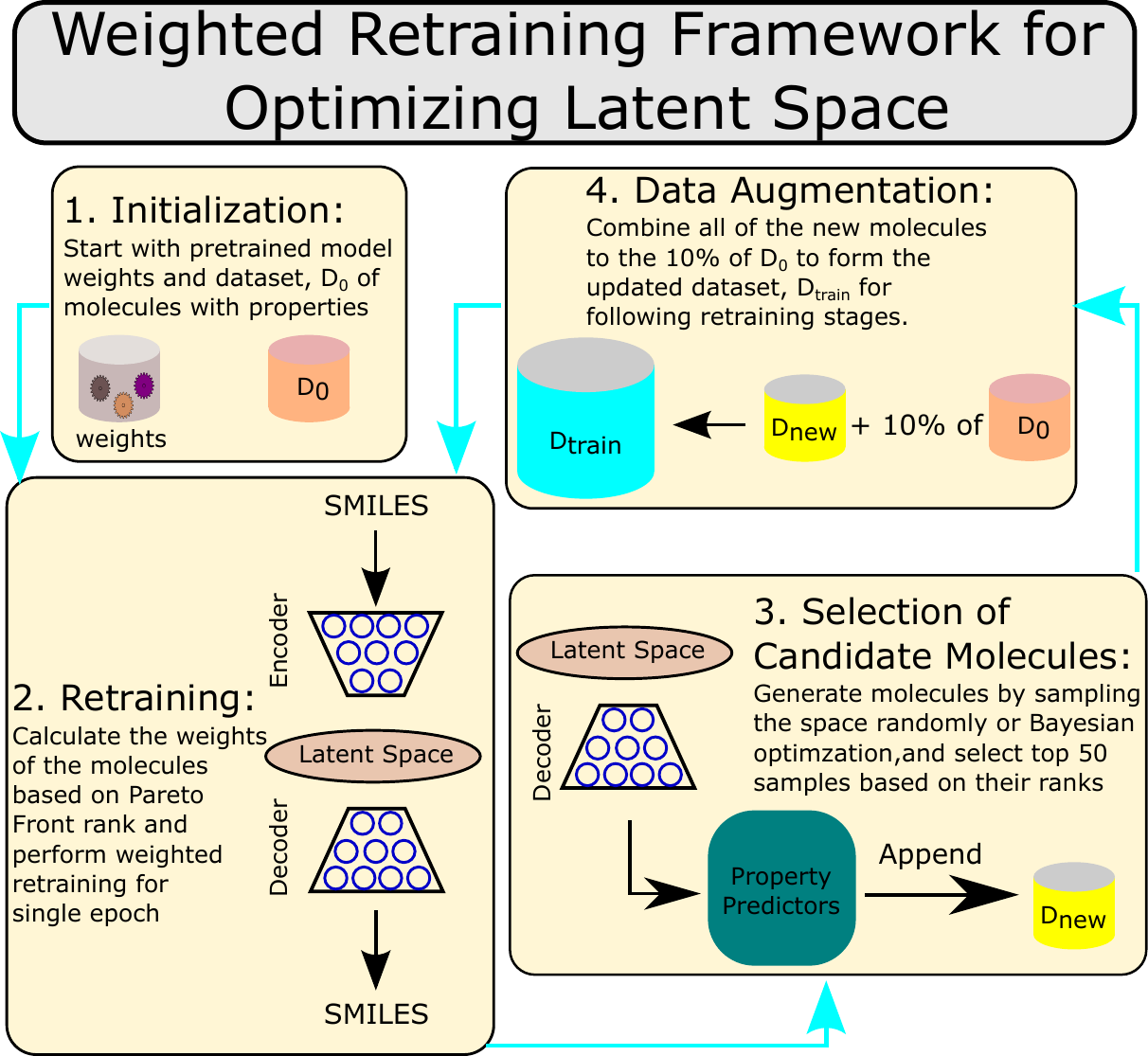}
    \caption{\textbf{Overview of the proposed multi-objective latent space optimization scheme.} The initial JT-VAE model is trained based on the original training dataset (Step-1). The weights of the molecules in the dataset are adjusted according to their Pareto front ranking based on the properties of interest. Desirable molecules with a higher ranking are assigned larger weights, while molecules with a lower ranking are assigned smaller weights. The JT-VAE is retrained based on the re-weighted dataset (Step-2). The retrained model is used to suggest novel molecules with enhanced properties by sampling or optimization in the latent space (Step-3). Top molecules are selected and used to augment the current training dataset (Step-4). Steps 2--4 may be repeated for iterative retraining of the generative model.}
    \label{fig:overview}
\end{figure}

In drug discovery applications, potential drug candidates always have to simultaneously meet multiple design criteria. For example, in addition to its capability to intervene with specific biomolecular target or mechanism, potential drug molecules are assessed based on various physio-chemical properties that contribute to their bioavailability.
In the generative molecule design schemes discussed above, multi-objective optimization is typically handled via scalarization, which turns the problem into a simpler single-objective optimization problem. However, the weights for combining the multiple objective functions are often selected by the designer in an ad hoc manner, despite their importance in guiding the multi-objective optimization process. This may potentially lead to sub-optimal results, as the optimization process may be unintentionally dominated by a few objective functions, resulting in emphasizing certain attributes while ignoring others.
Various other approaches have been also proposed to address multiple design criteria, where a notable example is RaionaleRL recently proposed by Jin et al.~\cite{jin20b_multi}. In this work, a policy gradient is applied to fine-tune a VAE for generating molecules from incomplete subgraphs that correspond to multi-property rationales. RationaleRL utilizes a  property predictor to extract the subgraph rationale for each of the multiple target properties. From the set of extracted single property rationales, RationaleRL identifies the combined subgraph that meets the multiple property constraints and then uses the fine-tuned generative model to complete the molecular graph for the combined subgraph. However, a practical limitation of this approach is that the number of training samples that satisfy all properties of interest may be scarce, which exacerbates as the number of target design criteria increases. MArkov moleculaR Sampling (MARS)~\cite{mars} proposed to address the multi-objective drug discovery problem by formulating the molecular design process as an iterative graph editing process. For this purpose, MARS defines the target distribution by combining the scoring functions of multiple properties and adopts Markov chain Monte Carlo (MCMC) sampling to identify high-scoring molecular candidates. However, as the target distribution is obtained by taking either the sum or the product of the multiple scoring functions, it faces similar shortcomings as other scalarization-based approaches discussed earlier. 
To generate molecules with multiple target properties, Feng et.al~\cite{feng2023multiobjective} utilized a Langevin diffusion process as a stochastic generator of latent embeddings within a pre-trained autoencoder model. This stochastic generator, primarily governed by reference molecules selected based on multiple properties of interest, can be seen as an alternative to other optimization strategies in latent space.

In this paper, we propose a novel multi-objective latent space optimization (MO-LSO) scheme that can effectively address the aforementioned limitations of existing generative models for molecular design 
-- specifically, their limited capability for extrapolation in a multi-objective fashion beyond the molecular property space seen during training. 
We extend the weighted retraining framework recently proposed in Tripp et al.~\cite{tripp2020sampleefficient} to equip it with the inherent capability to enhance the efficiency of sampling novel molecules in the latent space that simultaneously improve multiple target properties. This is achieved by ranking the molecules based on their Pareto optimality through non-dominated sorting (NDS), where the rankings are used both for generating improved molecules based on multiple design criteria to augment the training data and for determining the weights of the molecules in the (augmented) training set based on the relative importance. Our proposed MO-LSO scheme can naturally balance the trade-offs among multiple properties without any ad hoc scalarization that may potentially bias the optimization results. Furthermore, as the molecules are assessed based on their Pareto efficiency,  our MO-LSO scheme scales very well computationally as the number of design criteria increases (empirically demonstrated up to three objectives in this study)  and it is naturally equipped with the capability to handle the optimization of properties that may be highly correlated (or even redundant) or conflict one another.
We show that our proposed MO-LSO scheme can effectively shift the latent space representation of the molecules based on multiple design criteria, thereby substantially enhancing the sampling efficiency of the generative model for suggesting novel molecules that simultaneously improve multiple properties. Furthermore, by applying it to the design of DRD2 inhibitors, we also demonstrate  through {\it in silico} analysis  that the generative model optimized by the proposed MO-LSO scheme is able to produce highly promising molecules that outperform  known DRD2 inhibitory molecules. 

\section*{Results}

\subsection*{Overview of the proposed method and the experimental set-up}

Figure~\ref{fig:overview} provides an overview of our proposed multi-objective latent space optimization scheme for generative molecular design. Given an initial training dataset of molecules, we rank the molecules based on the multiple molecular properties of interest using the Pareto ranking scheme described in \eqref{eq:rank_eqn}. A detailed description of the algorithm for multi-objective ranking of the molecules can be found in the \textbf{Methods} section. To fine-tune the generative model's latent space to make it more sampling efficient for desirable molecules, the molecules in the dataset are weighted according to \eqref{eq:weight}, where higher-ranked molecules are assigned with larger weights while lower-ranked molecules are assigned with smaller weights. Retraining the generative model based on this weighted dataset biases the model toward higher-ranked molecules with more desirable properties, thereby making the latent space of the retrained model more amenable to suggesting novel molecules that are likely to be highly ranked based on the Pareto ranking scheme. 
After retraining the baseline model using the weighted-dataset for a single epoch, we explore the latent space to search for new molecules that can potentially improve upon the molecules in the dataset in terms of the multiple target properties. While various multi-objective optimization schemes may be adopted for this purpose, we considered two potential approaches in this current study: (i) random sampling and selection of the top-ranked molecules and (ii) single objective Bayesian optimization (SOBO). In the first approach, we generated $250$ random molecules out of which we selected the top $50$ molecules (based on Pareto ranking). In the second approach, we used SOBO \cite{frazier2018tutorial} to generate $50$ molecules, out of which the unique molecules were selected. To evaluate and rank the novel molecules, the latent points were first decoded into the original molecular space, where their properties were assessed. The selected top novel molecules
were then used to form a ``candidate dataset'' that can be used to augment the training dataset at hand, potentially pushing the current Pareto frontier and further improving the sampling efficiency of the current model for improved molecules.

Before the next retraining iteration, we create an updated training set by combining the selected top candidates with $10\%$ of the initial dataset, which consists of randomly selected molecules. The random downselection of the molecules in the initial training data mainly aims at reducing the computational cost needed for shifting the latent space toward desirable directions within fewer iterations. The updated training set is then used for the weighted retraining of the generative model, which is subsequently used for identifying a new set of desirable molecules to be appended to the candidate dataset. This \textit{iterative retraining cycle} -- new candidate generation based on the current model, augmentation of the candidate set and creation of a new training set that integrates the additional candidates, and re-ranking the molecules and performing another weighted retraining of the model -- can be repeated until either the generated molecules meet the desired multi-objective criteria, converge in terms of the molecular properties, or the total training cost (computation or time) reaches a prespecified budget. We further describe the iterative retraining procedure in the later subsections based on specific molecular optimization scenarios.

In this study, we considered the pairwise optimization of the following molecular properties: (i) water-octanol partition coefficient (logP), (ii) synthetic accessibility score (SAS), (iii) natural product-likeness score (NP score)~\cite{NP_score_paper}, and (iv) the probability of inhibition against the dopamine receptor D$_2$ (DRD2) \cite{creese1976dopamine}. We aimed to maximize logP, NP score, and DRD2 inhibition property. On the other hand, we aimed to minimize SAS, as a lower SAS indicates better synthesizability of a given molecule. For the computation of logP and SAS, the RDKit package~\cite{rdkit} was used. We adopted the method  in~\cite{NP_score_paper} for assessing the NP score. The probability of inhibition against DRD2 was estimated by using an ML surrogate model, whose details are given in the \textbf{Methods} section.


\subsection*{Weighted retraining via Pareto front rank effectively shifts the latent space for multiple objectives}

As the baseline model, we used the pretrained JT-VAE shared by Tripp et al.~\cite{tripp2020sampleefficient} and applied the proposed multi-objective weighted retraining scheme. Initially, we used the complete ZINC dataset\cite{zinc_dataset} for weighted training of the baseline model, where the dataset was split into training ($218969$ molecules) and validation sets ($24333$ molecules)  as in~\cite{tripp2020sampleefficient}. After each weighted retraining step, $250$ new molecules were randomly sampled, out of which the top $r=50$ candidates were selected based on the Pareto Front rank. The selected top candidates were used in the subsequent retraining stages. For details, please see the \textbf{Methods} section.

\begin{figure}[ht!]
    \centering
    \begin{minipage}{0.95\textwidth}
    \centering
    \includegraphics[width=0.9\textwidth]{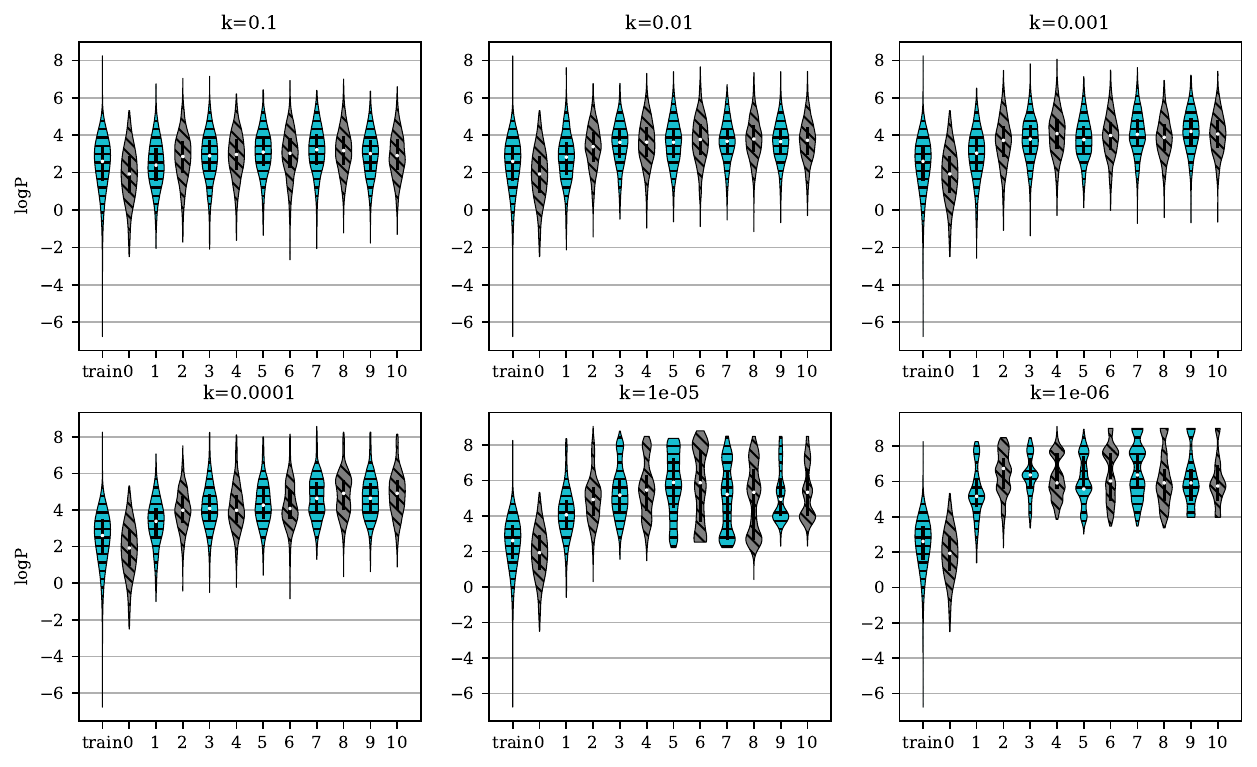}
    \end{minipage}
    \begin{minipage}{.95\textwidth}
    \centering
    \includegraphics[width=0.9\textwidth]{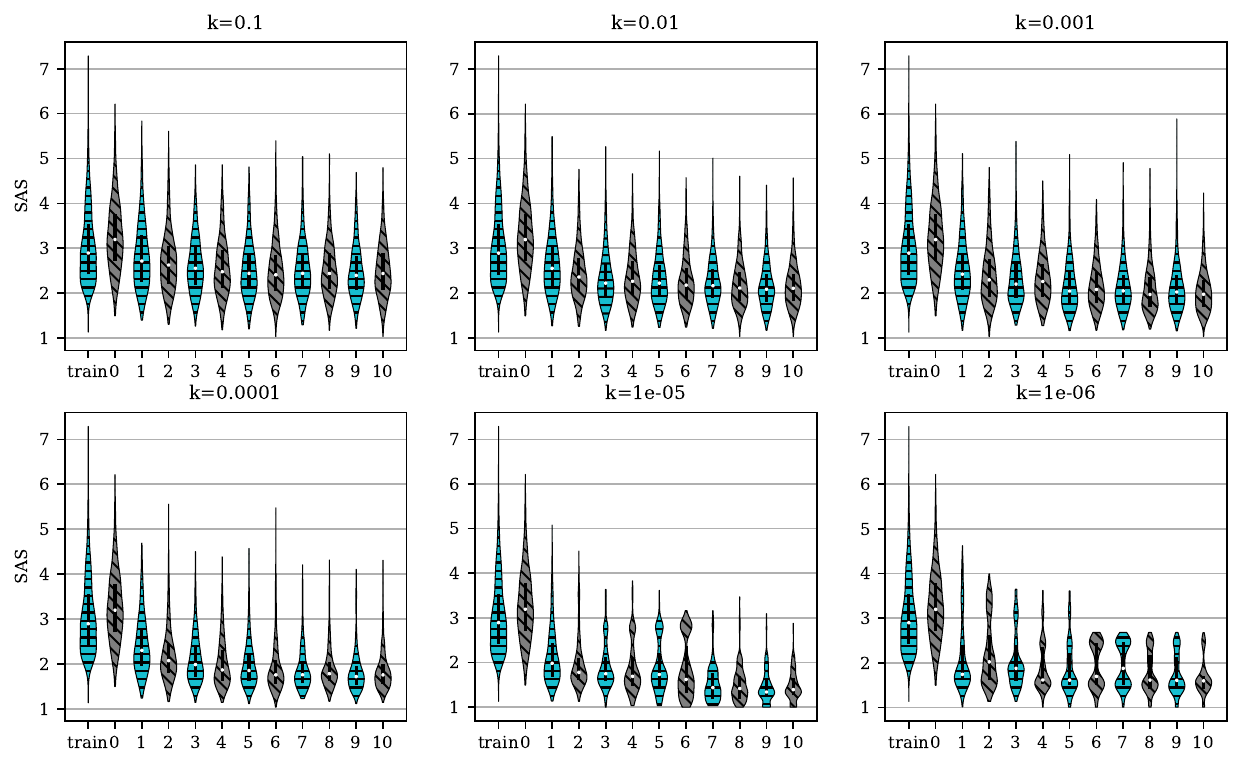}
    \end{minipage}
    \caption{\textbf{Evolution of the property distribution of the generated molecules due to latent space optimization via iterative weighted retraining.} The plots show how the property distribution changes as a result of weighted retraining of the JT-VAE based on the proposed multi-objective latent space optimization scheme. The latent space of the JT-VAE was optimized to suggest molecules with larger logP and smaller SAS. Results are shown for different values of $k$, which determines the sensitivity of the weight to ranking. In each subfigure, the first violin plot (labeled ``train'') shows the property distribution of all molecules in the initial training dataset. The subsequent violin plots show the property distribution of 1,000 randomly sampled molecules after $i$-th iterative retraining ($i=0$ corresponds to the original JT-VAE without any retraining). The results clearly show that the distribution of logP is shifted upward while that of SAS is shifted downward during the iterative retraining process as desired.}
    \label{logP_SAS_hist}
\end{figure}

Figure~\ref{logP_SAS_hist} shows the evolution of the property distribution of the generated molecules as we iterate the weighted retraining cycles. Each subplot shows the distributional changes over multiple iterations for a specific value of $k$, the hyper-parameter that determines how the rankings translate into the weights. Results are shown for optimizing the latent space of the JT-VAE for enhancing the property pair (logP, SAS). Here we retrained the baseline model $10$ times for several different values of $k$ ranging from $0.1$ to $10^{-6}$. Based on a given model at a specific iteration in the retraining cycle, we collected $1,000$ molecules randomly sampled from the latent space in order to plot the distribution of the molecular property of interest. The $x$-axis refers to the property distribution after the $i$-th iterative weighted retraining, where ``train'' corresponds to the property distribution of the molecules in the initial ZINC dataset. Furthermore, iteration 0 corresponds to the property distribution of the molecules sampled from the latent space of the baseline model (without any weighted retraining). These distributions are shown as a reference to show the relative improvement of the molecular property of interest as a result of the weighted retraining procedure.

As can be seen in Fig.~\ref{logP_SAS_hist}, the logP distribution of the molecules sampled from the latent space tends to shift upwards as desired as the retraining cycles proceed. Similarly, the SAS values tends to decrease as desired, indicating that iterative retraining generally improves the overall synthesizability of the molecules suggested by the JT-VAE. As discussed in \textbf{Methods}, a smaller $k$ places a greater emphasis on higher-scoring molecules. We can see its impact on the weighted retraining results in Fig.~\ref{logP_SAS_hist}, where using a smaller $k$ leads to a more rapid and more pronounced shift of the property distribution. However, the use of a smaller $k$ value makes the overall retraining process dominated by a smaller set of high-scoring molecules, which may have an impact on the overall diversity of the generated molecules and skew the property distribution of the molecules. For example, using $k = 10^{-5}$ or $k=10^{-6}$ results in bi-modal (or multi-modal) property distributions after several iterations of weighted retraining, reflecting this phenomenon. Here, the model learns the latent space mainly based on a limited number of high-scoring molecules at the Pareto front. Consequently, the molecules sampled in the learned latent space become clustered around those high-ranking molecules, which may limit the diversity of the molecules generated by the retrained model.

Table~\ref{A_tab_1} in the \textbf{Supplementary Information} shows the average property values for the molecules in the training data, molecules generated using the initial model (based on $1000$ randomly generated molecules), and molecules generated by the retrained model (again based on $1000$ randomly molecules) for different values of $k$. Additionally, we evaluated the structural diversity  (Table~\ref{A_tab_1})  of the molecules generated by different versions of JT-VAE models, in order to assess the ability of a given model to learn, represent, and sample from a wider chemical space.

The diversity was measured in terms of the average structural distance (based on ECFC4 fingerprints) over all pairs in a given set of molecules. For smaller values of $k$, we observe that the structural diversity is reduced as expected. As mentioned earlier, using a smaller $k$ assigns relatively higher weights to a smaller group of high-ranking molecules, which has the effect of making the model ``see'' this small group of molecules more frequently while retraining. Consequently, the model learns the latent space representation of the chemical space mainly based on these select molecules, which may make the molecules sampled from the learned latent space bear higher similarity to one another.

To further demonstrate the effectiveness of our proposed approach MO-LSO, we performed additional experiments for simultaneous optimization of two or three molecular properties and compared the results against the scalarization baseline of \cite{tripp2020sampleefficient} 
and a Markov molecular sampling scheme for multi-objective drug discovery, called MARS \cite{mars}.
Specifically,  the latter approach involves training a molecular generative model from scratch targeting multiple properties of interest whereas our approach starts with a generative model that is trained in a self-supervised fashion without the molecular property values of the training samples. However, through iterative retraining via our approach, the generative model succeeds in outperforming MARS for quite a few property combinations (Table \ref{mars_exp}).
Furthermore, we investigated the impact of retraining on the JT-VAE's molecule reconstruction performance.
The results obtained from these experiments are summarized and discussed in \textbf{Supplementary Information}.

\subsection*{Multi-objective latent space optimization effectively recovers sampling efficiency for incomplete dataset}

\begin{figure}[ht!]
    \centering
    \begin{minipage}{.2\textwidth}
    \centering
    \includegraphics[width=\textwidth]{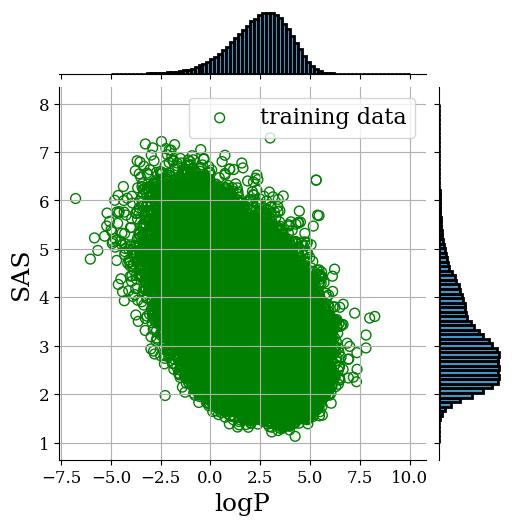}
    \end{minipage}%
    \begin{minipage}{.2\textwidth}
    \centering
    \includegraphics[width=\textwidth]{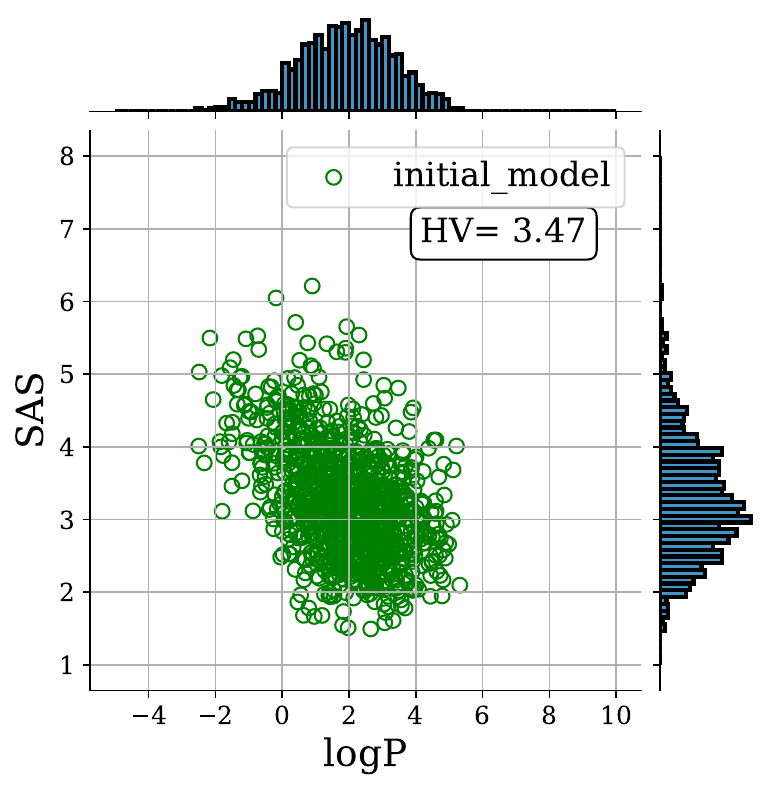}
    \end{minipage}%
    \begin{minipage}{.2\textwidth}
    \centering
    \includegraphics[width=\textwidth]{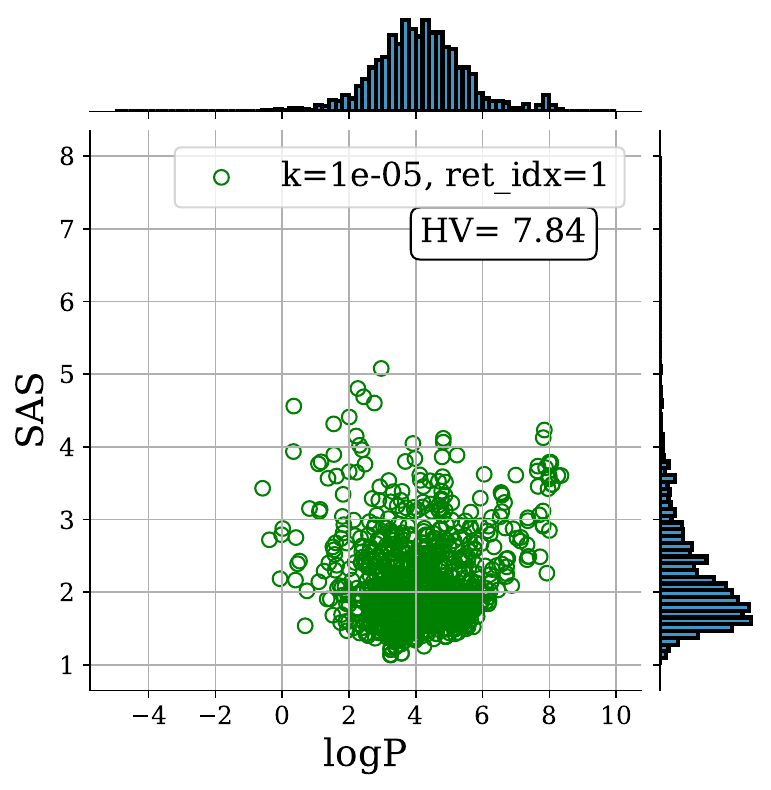}
    \end{minipage}%
    \begin{minipage}{.2\textwidth}
    \centering
    \includegraphics[width=\textwidth]{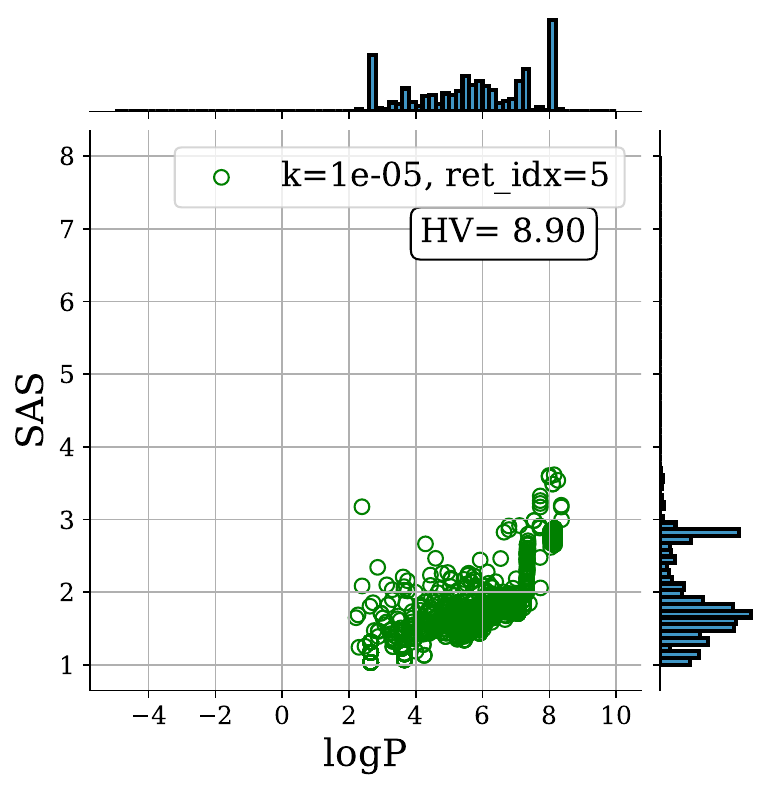}
    \end{minipage}%
    \begin{minipage}{.2\textwidth}
    \centering
    \includegraphics[width=\textwidth]{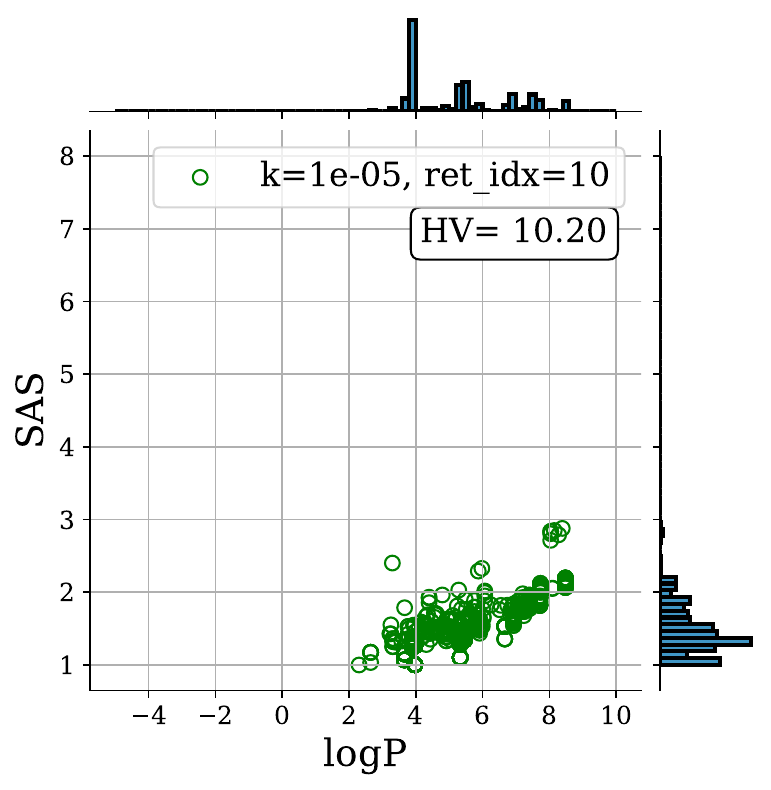}
    \end{minipage}
    \begin{minipage}{.2\textwidth}
    \centering
    \includegraphics[width=\textwidth]{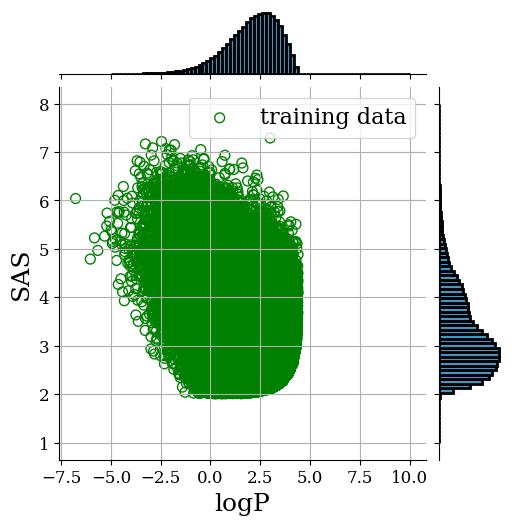}
    \end{minipage}%
    \begin{minipage}{.2\textwidth}
    \centering
    \includegraphics[width=\textwidth]{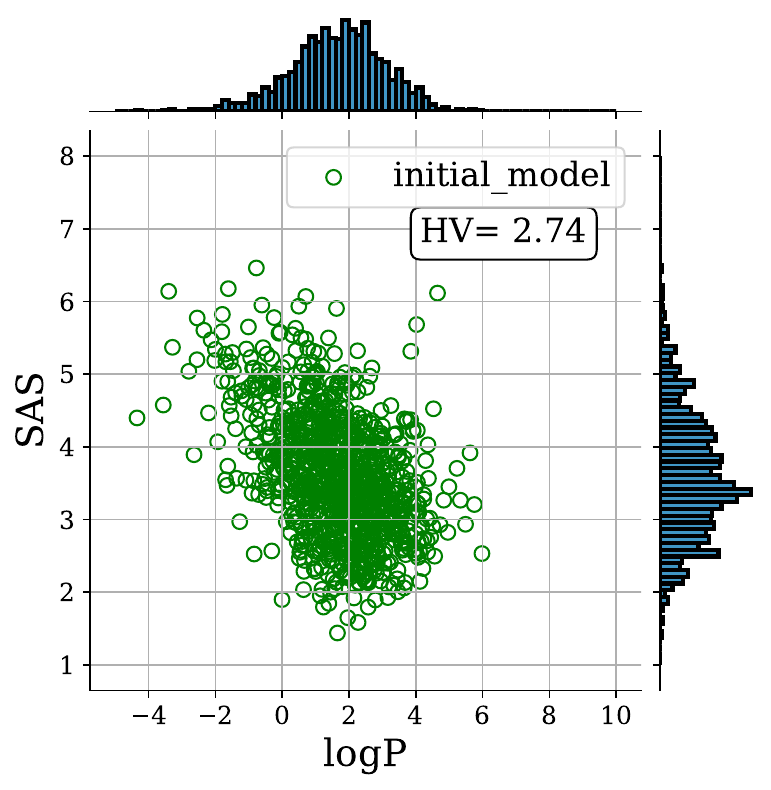}
    \end{minipage}%
    \begin{minipage}{.2\textwidth}
    \centering
    \includegraphics[width=\textwidth]{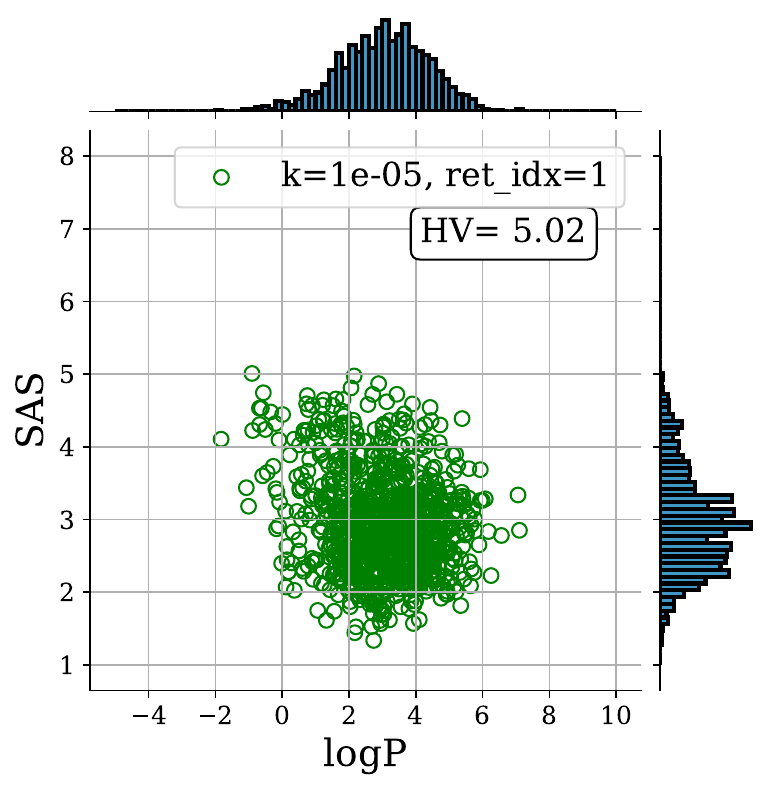}
    \end{minipage}%
    \begin{minipage}{.2\textwidth}
    \centering
    \includegraphics[width=\textwidth]{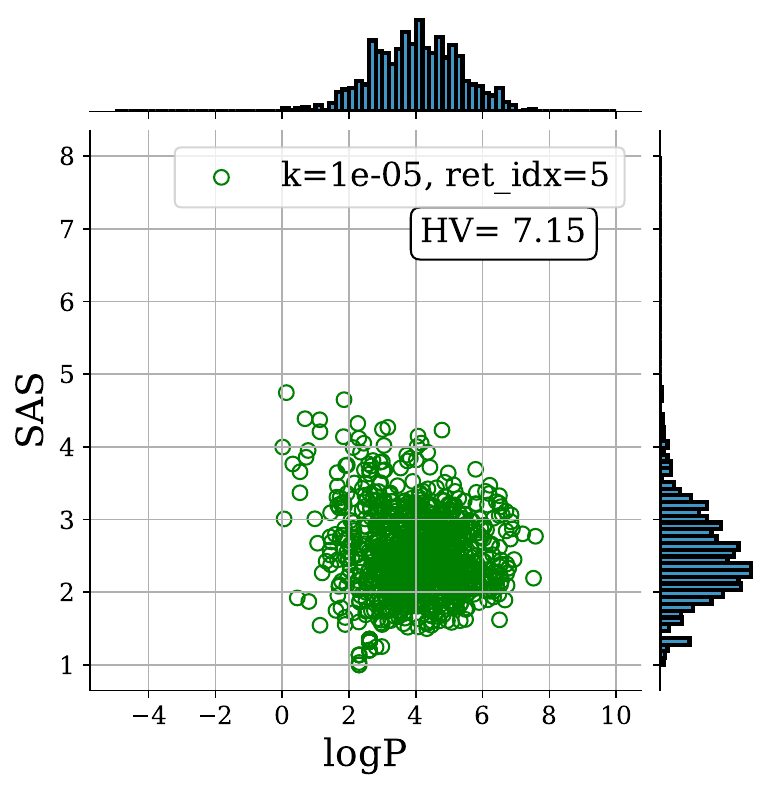}
    \end{minipage}%
    \begin{minipage}{.2\textwidth}
    \centering
    \includegraphics[width=\textwidth]{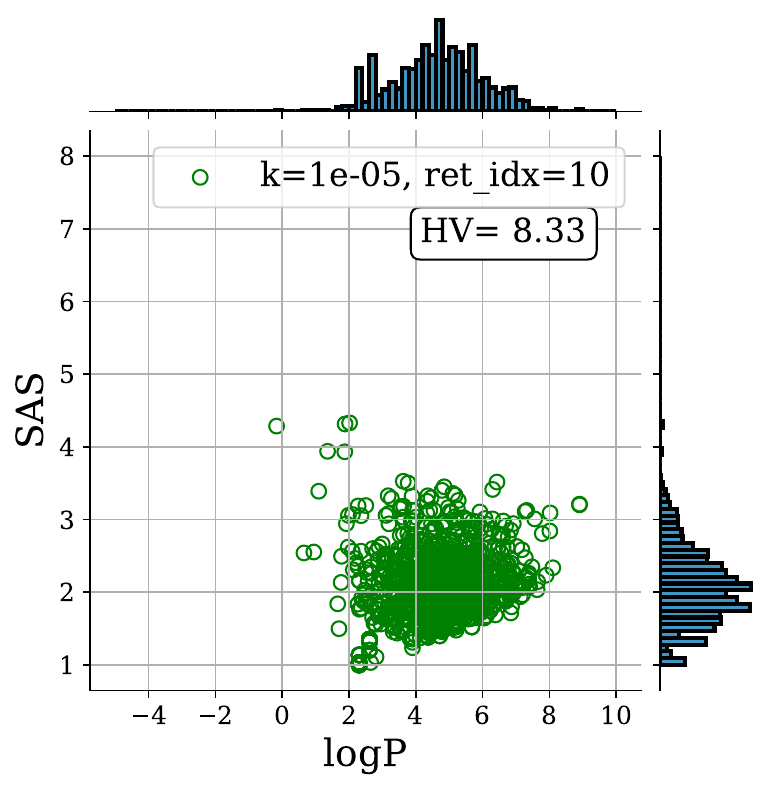}
    \end{minipage}
    
    \caption{\textbf{Evolution of the Pareto front via latent space optimization of the generative model.} The latent space of the JT-VAE has been jointly optimized to maximize logP and minimize SAS of the molecules suggested by the generative model. The scatter plots show the (logP, SAS) distribution of the molecules in the initial training dataset (column 1), molecules sampled in the latent space of the baseline model (column 2),  and molecules suggested by the optimized model after iteration-1 (column 3), iteration-5 (column 4), iteration-10 (column 5). The plots in the top row show the trends for the case when the complete training dataset was used, while the bottom row shows the trend when a reduced dataset was used. The results show that the Pareto front gradually shifts towards the desired direction (i.e., bottom right for larger logP and smaller SAS) resulting in a larger hypervolume (HV) of the Pareto front dominated property space in both cases.}
    \label{scatter_t1_t2}
\end{figure}

Next, we investigated the ability of the weighted retraining scheme to propose high-scoring molecules when such molecules are absent in the training dataset.
For this purpose, we first removed the top $20\%$ molecules -- selected based on their Pareto front rank -- from the training data that is used to train the baseline model~\cite{tripp2020sampleefficient}. Since the original baseline model has seen the complete dataset during its training, for each property pair, we trained a separate baseline model based on the reduced dataset that does not contain the top 20\% molecules with the highest Pareto front rank for the given property pair. In all cases, a learning rate of $0.0007$ was used with a batch size of $32$ for $30$ epochs for model training.
The scatter plots in Fig.~\ref{scatter_t1_t2} show the progression of the two properties logP and SAS, where the latent space of the JT-VAE is optimized for the given property pair with $k=10^{-5}$.

In each plot, we also show the hypervolume of the property space that is dominated by the Pareto front. The hypervolume is computed with respect to the average property values of the molecules in the complete training dataset. 
The top row of Fig.~\ref{scatter_t1_t2} shows the evolution of the Pareto front when the complete dataset was used. On the other hand, the bottom row in Fig.~\ref{scatter_t1_t2} shows the trends for the reduced dataset which does not contain the top 20\% molecules. As can be seen in the bottom row of Fig.~\ref{scatter_t1_t2}, although the initial dataset does not contain many molecules with logP greater than $4$ and SAS lower than $2$, the iterative multi-objective weighted retraining still manages to effectively push the latent space towards a desirable region that contains high-scoring molecules, where the trends are similar to the case when the complete dataset is used.  This is also illustrated by the larger hypervolume achieved by the optimized model compared to the initial pre-trained model in both cases.


The capability of the proposed multi-objective LSO method to recover high-performance molecules - despite the absence of such molecules in the training data - is demonstrated even more clearly when applied to the optimization of inhibitory molecules for DRD2. In this experiment, we considered pairwise property optimization of DRD2 inhibition along with one of the properties among logP, SAS, and NP score. As before, in each experiment for a given property pair, we removed the top $20\%$ molecules from the training data based on the Pareto Front rank. What makes this experiment especially interesting is the fact that the training dataset is highly imbalanced and contains a relatively small number of inhibitory molecules against DRD2. As a result, the removal of the top $20\%$ molecules leaves virtually no active DRD2 inhibitors in the training data. Moreover, since we train a new baseline model with the reduced dataset, the trained model does not initially possess any knowledge of inhibitory molecules against DRD2. 
Consequently, the multi-objective LSO method needs to guide the optimization of the latent space of the generative model towards a completely unexplored region, making the optimization task more challenging.

%

\begin{figure}[ht!]
    \centering
    \begin{minipage}{.2\textwidth}
    \centering
    \includegraphics[width=\textwidth]{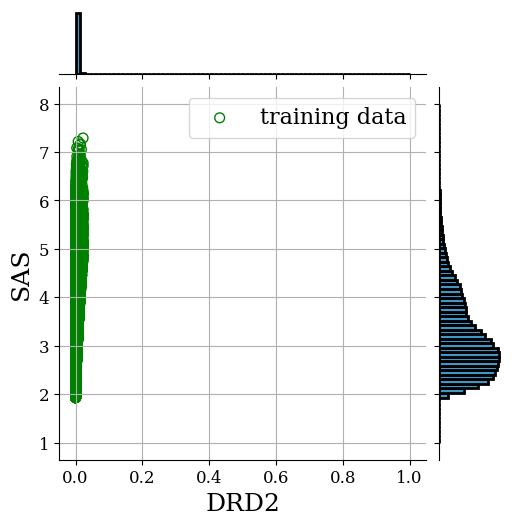}
    \end{minipage}%
    \begin{minipage}{.2\textwidth}
    \centering
    \includegraphics[width=\textwidth]{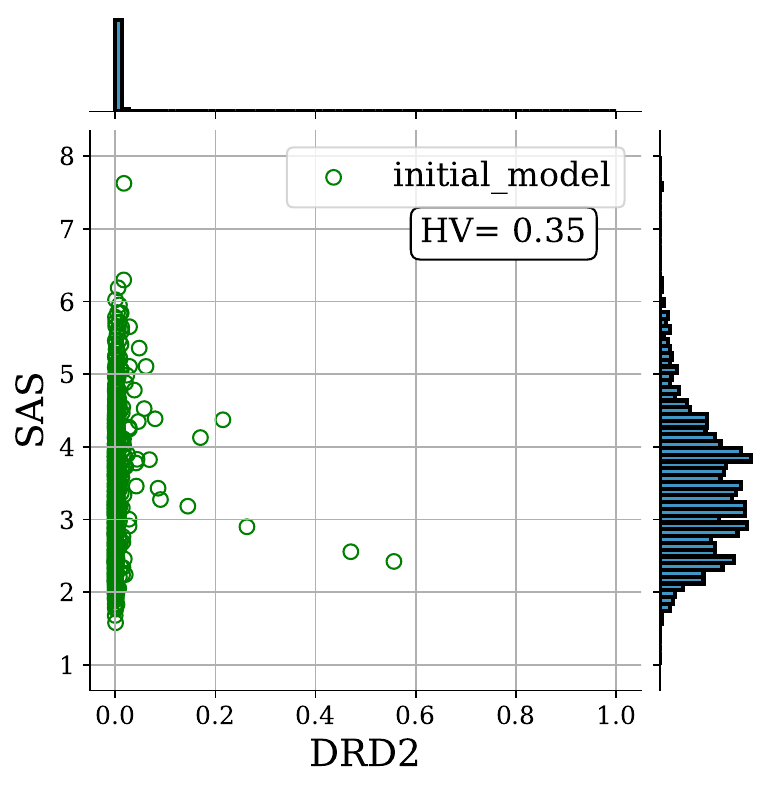}
    \end{minipage}%
    \begin{minipage}{.2\textwidth}
    \centering
    \includegraphics[width=\textwidth]{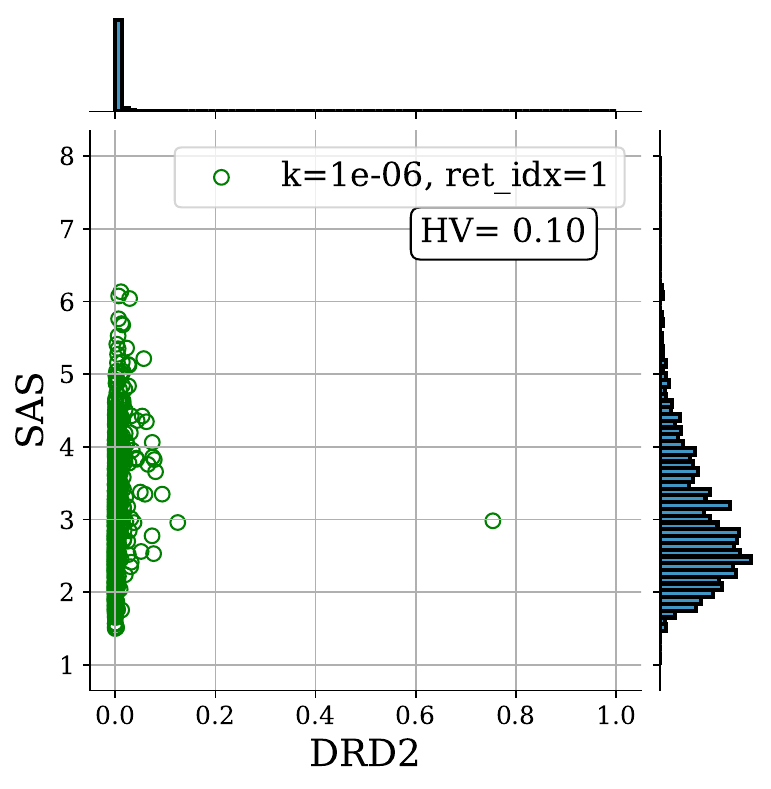}
    \end{minipage}%
    \begin{minipage}{.2\textwidth}
    \centering
    \includegraphics[width=\textwidth]{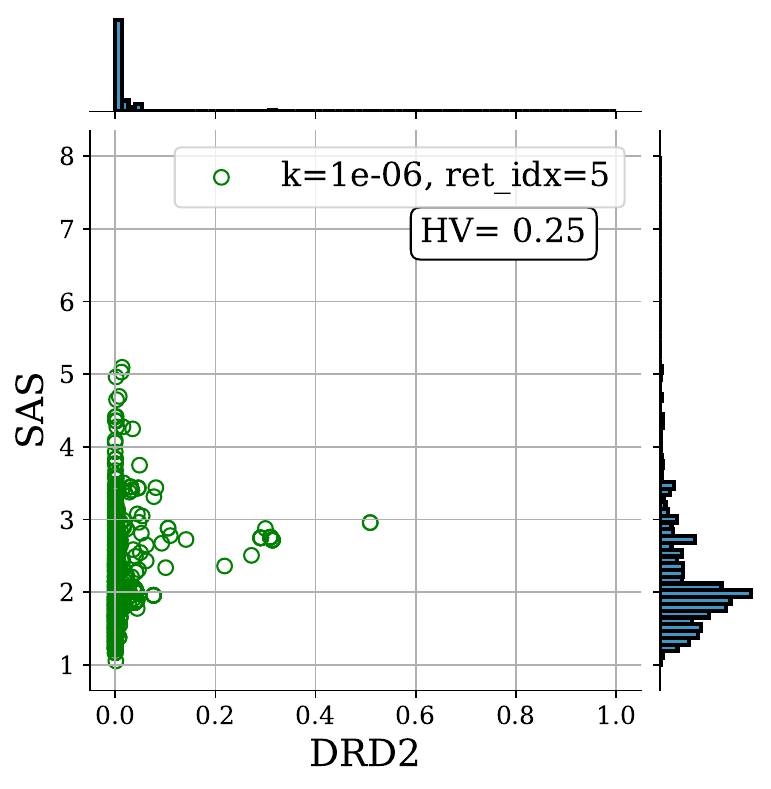}
    \end{minipage}%
    \begin{minipage}{.2\textwidth}
    \centering
    \includegraphics[width=\textwidth]{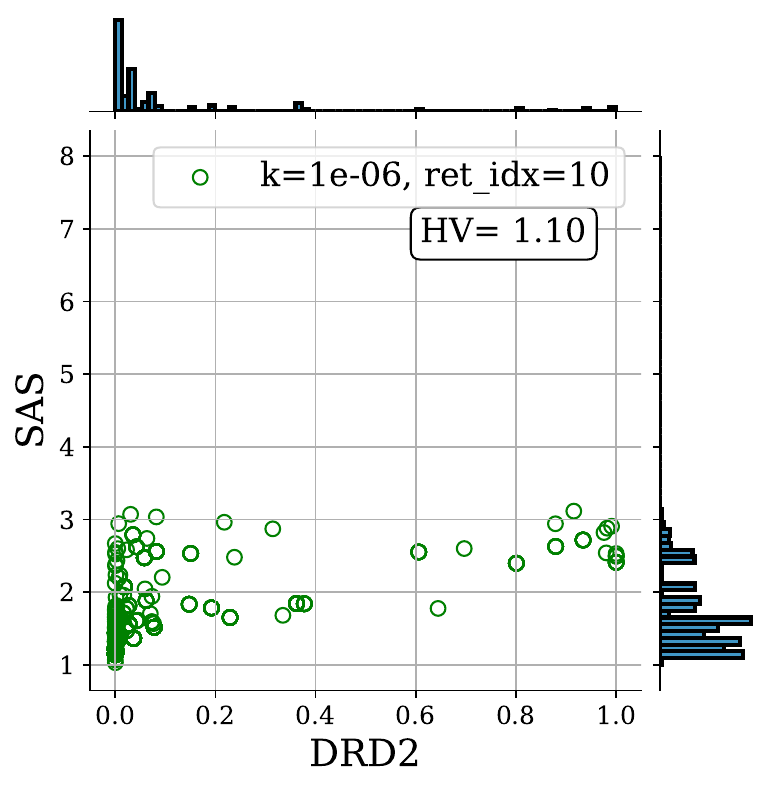}
    \end{minipage}

    \caption{\textbf{Transition of DRD2 and SAS towards optimum direction while starting with no DRD2 active training samples.} The first two scatter plots represent the training data and the molecules generated from the baseline model respectively. In rest of the jointplots, the molecules from learned model after $1^{st}$, $5^{th}$ and $10^{th}$ weighted retraining for $k=10^{-6}$ are shown in the objective space. The hypervolume (HV) in each plot indicates the volume of the property space that is dominated by the Pareto front. As we iterate the weighted retraining, the resulting hypervolume tends to increase.}
    \label{scatter_drd2_sas}
\end{figure}

Figure~\ref{scatter_drd2_sas} shows the results for the multi-objective weighted retraining with $k=10^{-6}$ for the property pair DRD2 and SAS. The leftmost scatter plot shows the distribution of the training data after removing the top 20\% molecules. The horizontal axis shows the probability of inhibition (the higher the better) and the vertical axis shows the SAS (a lower score corresponds to better synthesizability). The second subplot (from the left) shows the property distribution of $1,000$ molecules randomly sampled in the latent space of the baseline model. Next, the third, fourth, and fifth (i.e. rightmost) subplots depict the property distribution of the molecules after the $1^{st}$, $5^{th}$ and $10^{th}$ weighted retraining cycle, respectively. While the initial training data are absent of DRD2 inhibitory molecules, we can see that a relatively large number of inhibitory molecules after the $10^{th}$ weighted retraining.
Moreover, the SAS distribution of the sampled molecules becomes more skewed towards smaller values as expected.
We have repeated similar experiments for two other property pairs (DRD2, logP) and (DRD2, NP score), which all showed similar trends. Details of these simulation results can be found in Table~\ref{A_tab_2} of \textbf{Supplementary Information}.

\subsection*{In silico analysis of the designed DRD2 inhibitors}

Given the long history of structure-based rational design, the use of methods that are structurally unaware to design ``active" molecules can be quite disconcerting. However, the use of the SVM model training with thousands of active and inactive compounds that sample the available contacts in the pocket seems to have encoded the space well enough that while the method is not structurally aware the information required is implicitly available. Potentially of more value is removing the inherent bias in what an ``active should look like'' seen when working with computational and medicinal chemists. In this study, tossing out the concept of what a hit should look like and taking a data-driven generative approach for molecular design led to some strange-looking molecules, that have attractive properties and by multiple measures of \textit{in silico} modeling (docking, MM-GBSA and long duration MD) were superior to known DRD2 inhibitors.
\begin{figure}[ht!]
    \centering
    \begin{minipage}{.47\textwidth}
    \centering
    \includegraphics[scale = 0.95]{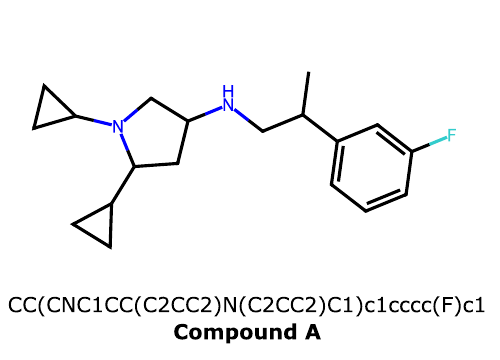}
    \caption*{(a)}
    \end{minipage}%
    \begin{minipage}{.43\textwidth}
    \centering
    \includegraphics[scale = 0.95]{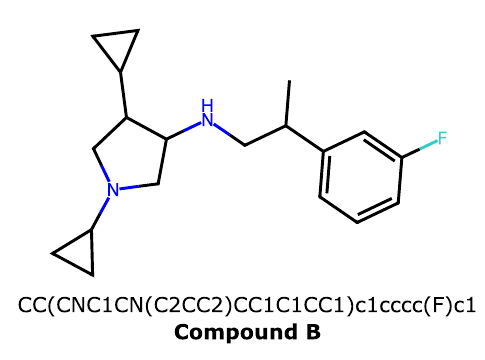}
    \caption*{(b) }
    \end{minipage}
    \caption{\textbf{Molecular structure and SMILES of two generated compounds with predicted DRD2 activity.} Both compounds A and B show the lowest docking energy among the pool of molecules which we generate by considering the DRD2 activity as one of the objectives. The weighted retraining framework starts from the initial training dataset that does not contain any active molecules.} 
    \label{drd2_compounds}
\end{figure}
For example, one of the designed compounds CC(CNC1CC(C2CC2)N(C2CC2)C1)c1cccc(F)c1 (which we refer to as ``compound~A'') have energies $-3.43kcal/mol$ lower than the crystal ligand by docking, $-3.78$ with MM-GBSA. Furthermore, molecular dynamics (MD) simulations showed stable, unstrained interactions over the length of the simulation.\\
An MD system was created using DPPC for the membrane, TIP4PEW water, and physiological levels of NaCl. Following $100pS$ of minimization the system was run using Desmond and the OPLS4 force field at $310.15K$ for $15nS$, and $100nS$, to evaluate the movement and ligand pose stability. For docking, Glide grids were created from the system at time $14.98$nS as well as the original crystal structure, and all residues within $5\angstrom$ of the ligand were set to allow rotation. Glide XP was used for docking since while computationally very expensive the sampling and pose minimization seem to best replicate the crystal position of the native ligand.

The generated novel molecules with predicted DRD2 activity were prepared using Ligprep in Schrodinger 2021.3, with the molecules enumerated around chiral centers in a pH range of $7.4 \pm 2.0$. The crystal structure of 6CM4 was prepared with the protein preparation tool in Schrodinger 2021.3 at a pH of $7.4$. 
The generated ligands were docked along with $667$ inactive bait compounds as well as two non-reverse agonist DRD2 ligands as controls due to the possibility that the Risperidone bound structure would not be favorable to the docking of traditional antagonists. This concern was unfounded since both Domperidone and L-741626 were able to achieve low energy poses. From the docking, two compounds - compound~A (shown in Fig. \ref{drd2_compounds}(a)) and another compound which we refer to as compound~B (shown in Fig. \ref{drd2_compounds}(b)) achieved the lowest energies of any compounds including the crystal ligand. MD simulations using the same parameters as before were run for $15nS$ on compounds A and B to evaluate whether the confirmations predicted were stable (was the compound ejected from the pocket) and if relevant contacts were maintained. Additionally, the Schrodinger Prime molecular mechanics generalized Born surface area (MM-GBSA) calculations was carried out on all compounds/poses with docking scores the same or equal to the lowest scoring pose of the crystal ligand.

\begin{figure}[t!]
    \centering
    \includegraphics[scale = 0.5]{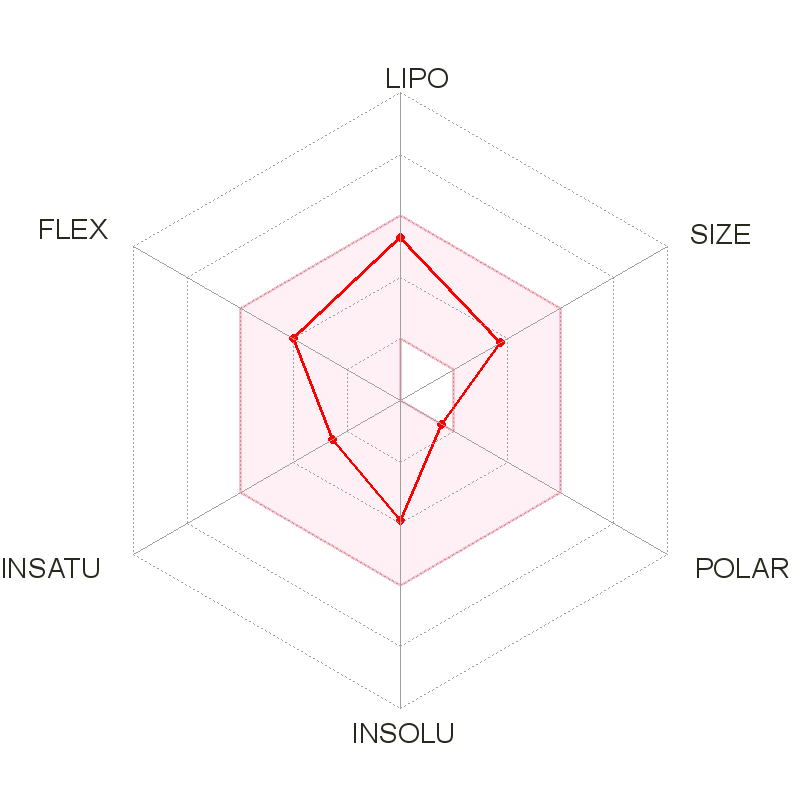}
    \caption{\textbf{Properties of a top molecule predicted by the optimized generative model.} The properties of a top molecule (compound~A) suggested by the JT-VAE, whose latent space was optimized by the proposed method. Six parameters -- POLAR (Polarity), INSOLU (Insolubility), INSATU (Instauration), FLEX (Rotable bond flexibility), LIPO (Lipophilicity), and SIZE (Molecular Weight) -- are shown. We can see that the suggested compound is within the colored zone, which corresponds to the physio-chemical space suitable for oral bioavailability \cite{SwissADME_paper}.}
    \label{comp7_radar}
\end{figure}

\begin{figure}[ht!]
    \centering
    \begin{minipage}{.45\textwidth}
    \centering
    \includegraphics[scale = .15]{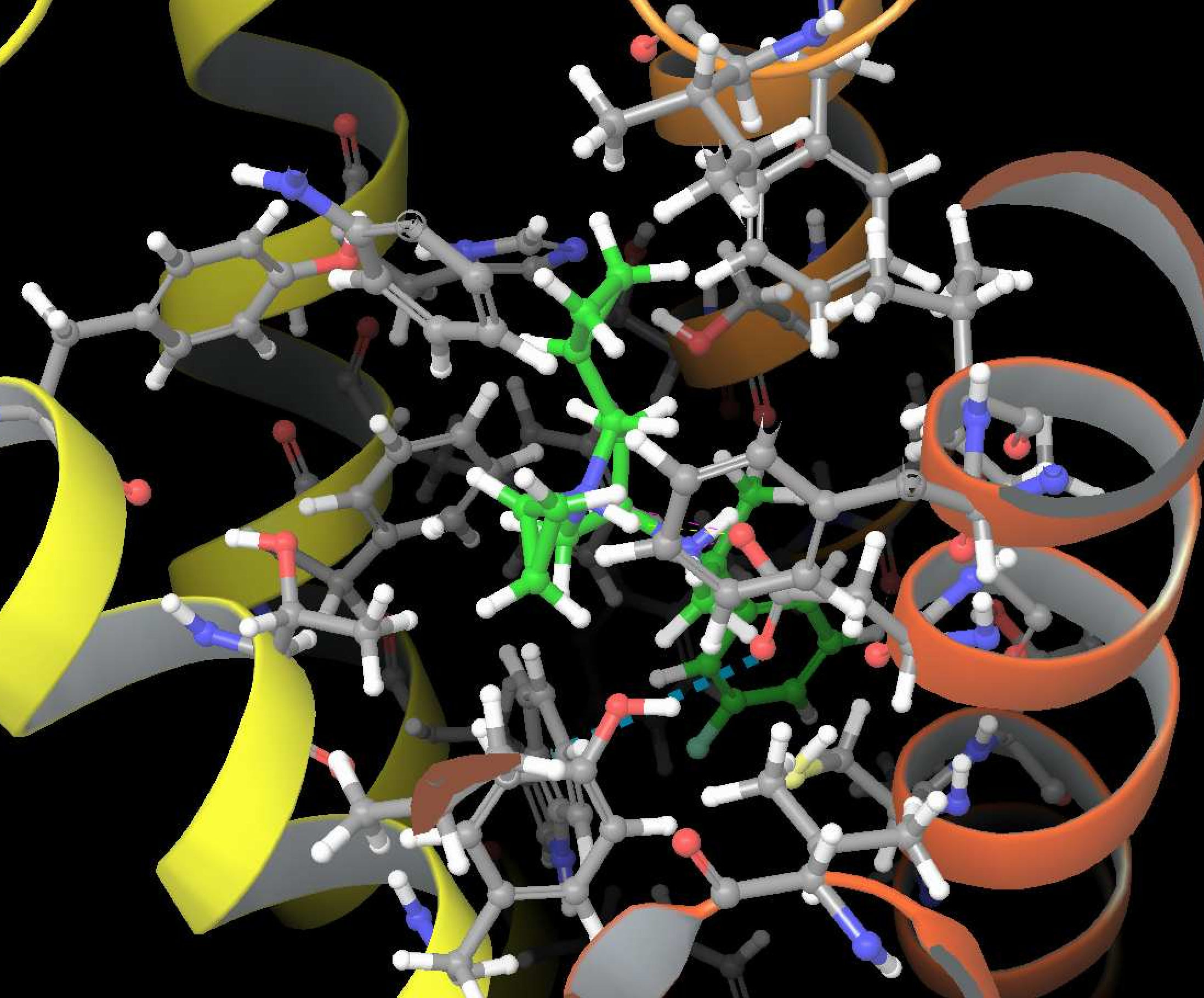}
    \caption*{(a)} 
    \end{minipage}%
    \begin{minipage}{.45\textwidth}
    \centering
    \includegraphics[scale = 0.165]{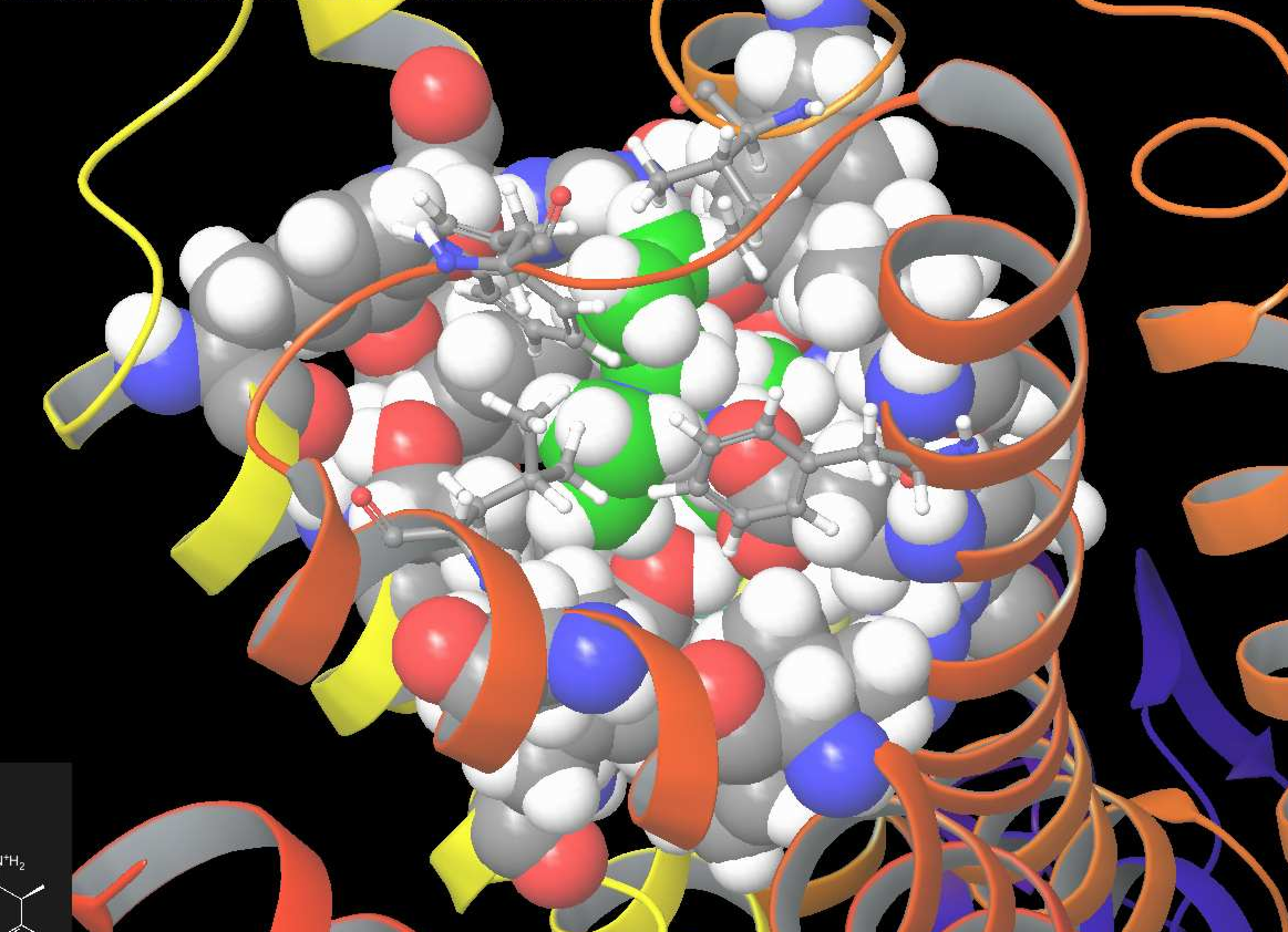}
    \caption*{(b) }
    \end{minipage}
    \caption{\textbf{Binding pose of compound~A in the DRD2 structure.} (a) The ball and stick representation shown highlights how the compound fits the broader pocket. (b) The space-fill model highlights the interaction between cyclopropanes and the hydrophobic mouth of the receptor.} 
    \label{md_plot_12}
\end{figure}

While \textit{in silico} results do not often translate in activity \textit{in vivo}, they are often used to select what molecules get made and their priority for testing, so the results of this effort and the computational chemistry evaluation produced ``high priority compounds'' that are very different from the native ligand and the molecules on the market. This ability to \textit{in silico} explore vast amounts of novel chemistry space while still generating compounds with desirable properties is likely to lead to the types of cost and time reductions in discovery that have been promised by the AI/ML community. The actual compounds predicted have desirable properties (Figure~\ref{comp7_radar} shows the properties for compound~A as predicted by SwissADME~\cite{SwissADME_paper}).
Additionally, the binding of compound~A (Figure~\ref{md_plot_12} (a)) shows that the typical aromatic interaction between W386 and the ligand in this case the $6$ member ring on the ligand as well as the potential for salt bridges from the ligand N's to the O's on D114. Of more interest is the two cyclopropane groups, which allow the compounds to have hydrophobic interactions with the largely hydrophobic mouth of the pocket, which includes residues 389, 392, 184, 189, 416, 412, etc. (Figure~\ref{md_plot_12} (b)) without having undesirable steric effects that bulkier ring systems may present. The potential desolvation effects from these residues while still allowing for the protein to have a closed confirmation have the potential to allow the compounds to be selective and active at low concentrations. As previously mentioned, this molecular series would not have been ranked high by computational or medicinal chemists based on how they looked, but once the poses were reviewed and the ease of the synthetic routes evaluated it is generally agreed that the series is a high priority.  For illustrative purpose, we performed the ADMET analysis for the top candidate for DRD2. In practice, if one desires to select the optimal candidate leads for the properties of interest, a more comprehensive screening like the one performed in \cite{feng2023multiobjective} may be required. 

\subsection*{Molecule optimization and selection via Bayesian optimization can enhance latent space optimization results}

\begin{figure}[ht!]
    \centering
    \begin{minipage}{.2\textwidth}
    \centering
    
    \includegraphics[width=\textwidth]{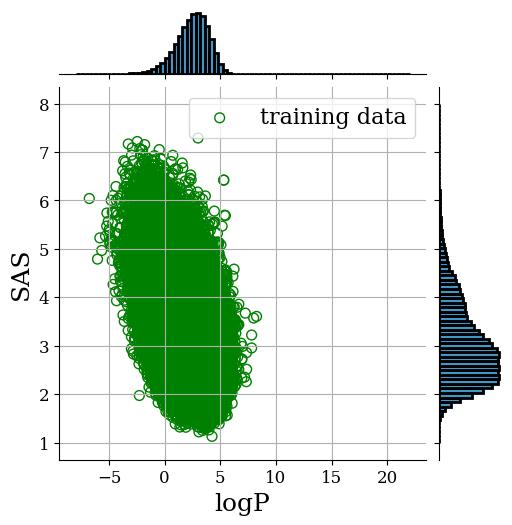}
    \end{minipage}%
    \begin{minipage}{.2\textwidth}
    \centering
    \includegraphics[width=\textwidth]{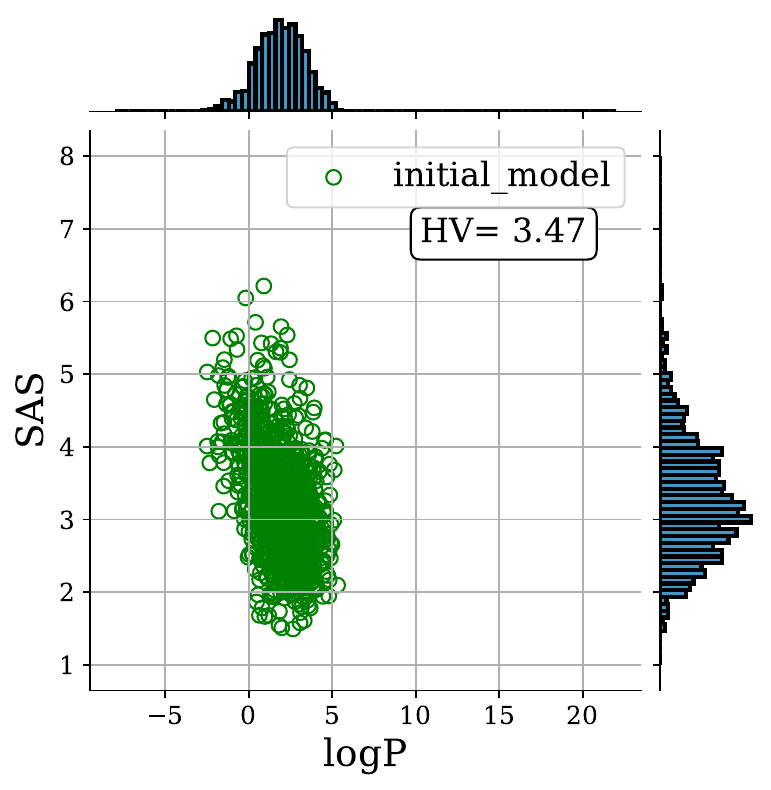}
    \end{minipage}%
    \begin{minipage}{.2\textwidth}
    \centering
    \includegraphics[width=\textwidth]{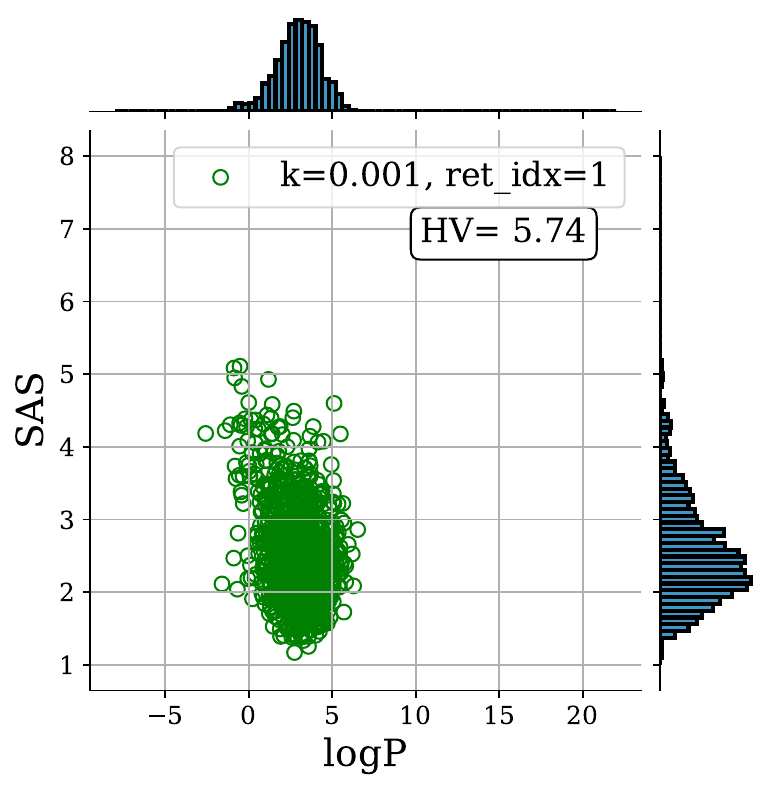}
    \end{minipage}%
    \begin{minipage}{.2\textwidth}
    \centering
    \includegraphics[width=\textwidth]{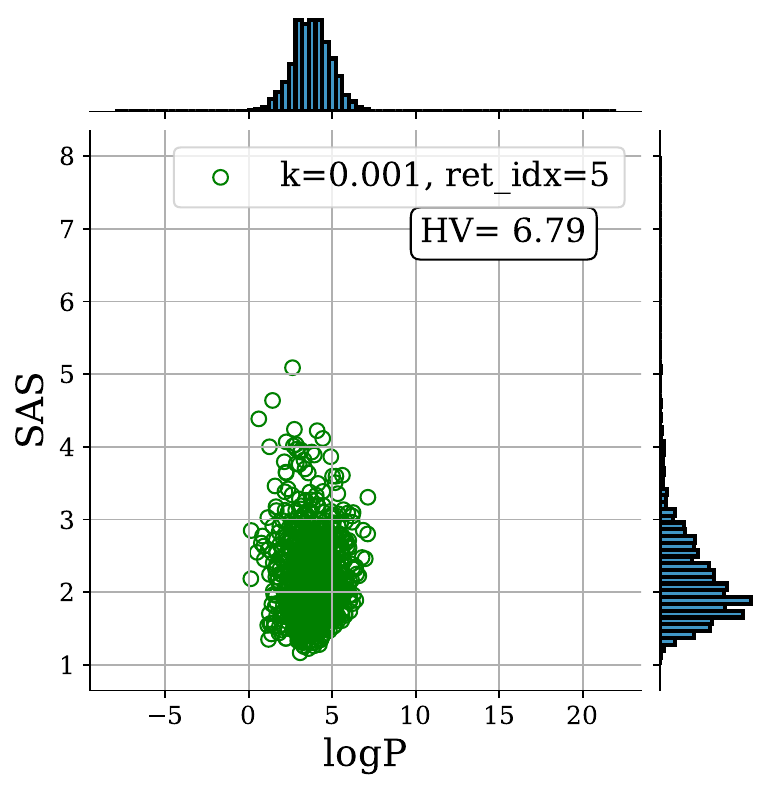}
    \end{minipage}%
    \begin{minipage}{.2\textwidth}
    \centering
    \includegraphics[width=\textwidth]{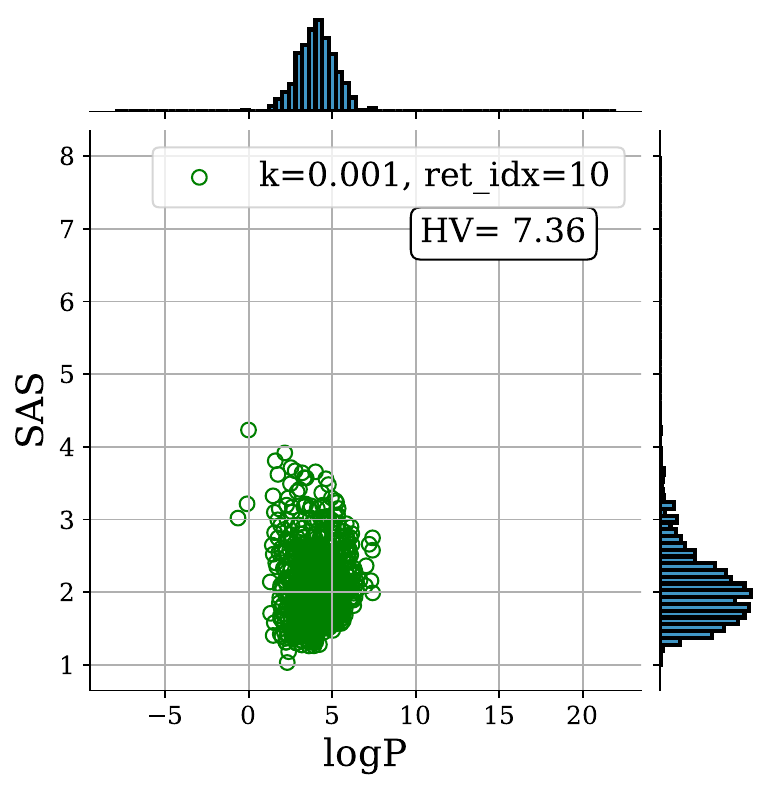}
    \end{minipage}
    
    \begin{minipage}{.2\textwidth}
    \centering
    
    \includegraphics[width=\textwidth]{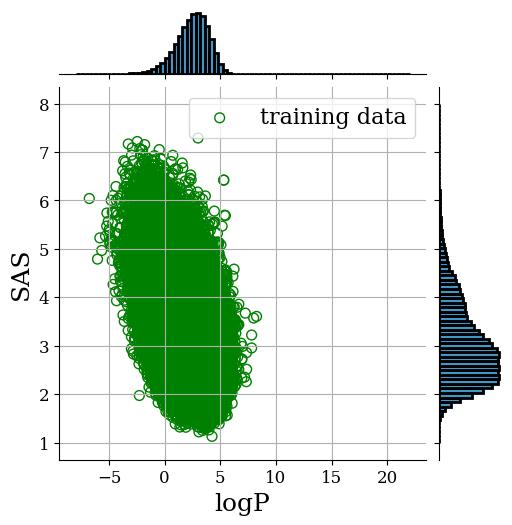}
    \end{minipage}%
    \begin{minipage}{.2\textwidth}
    \centering
    \includegraphics[width=\textwidth]{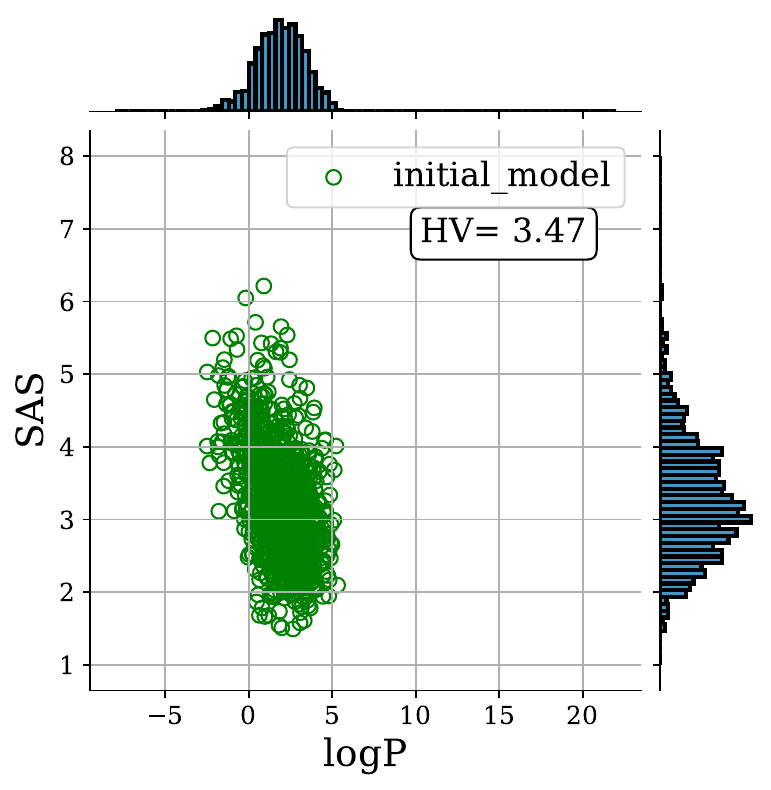}
    \end{minipage}%
    \begin{minipage}{.2\textwidth}
    \centering
    \includegraphics[width=\textwidth]{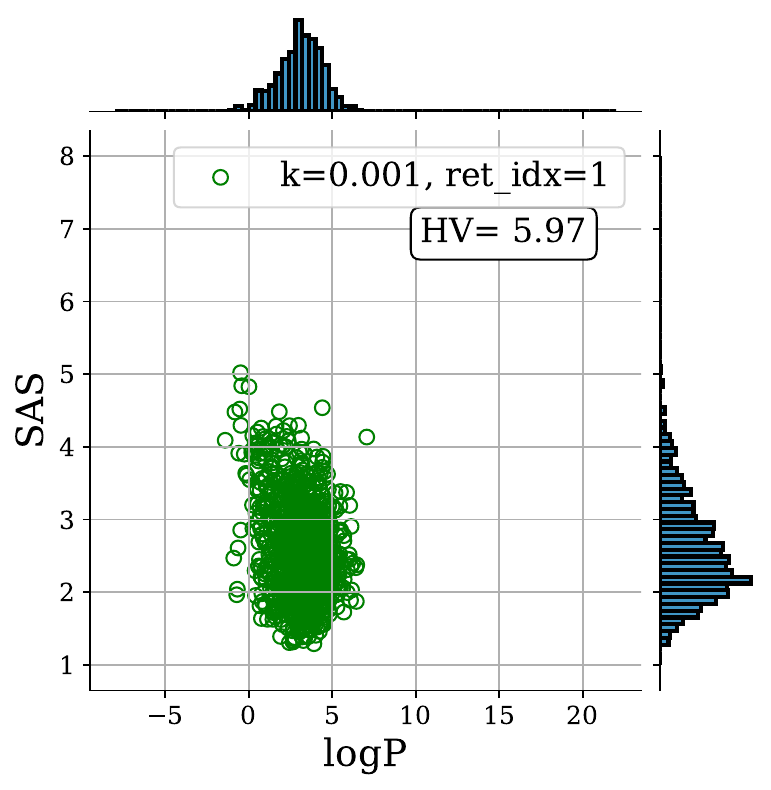}
    \end{minipage}%
    \begin{minipage}{.2\textwidth}
    \centering
    \includegraphics[width=\textwidth]{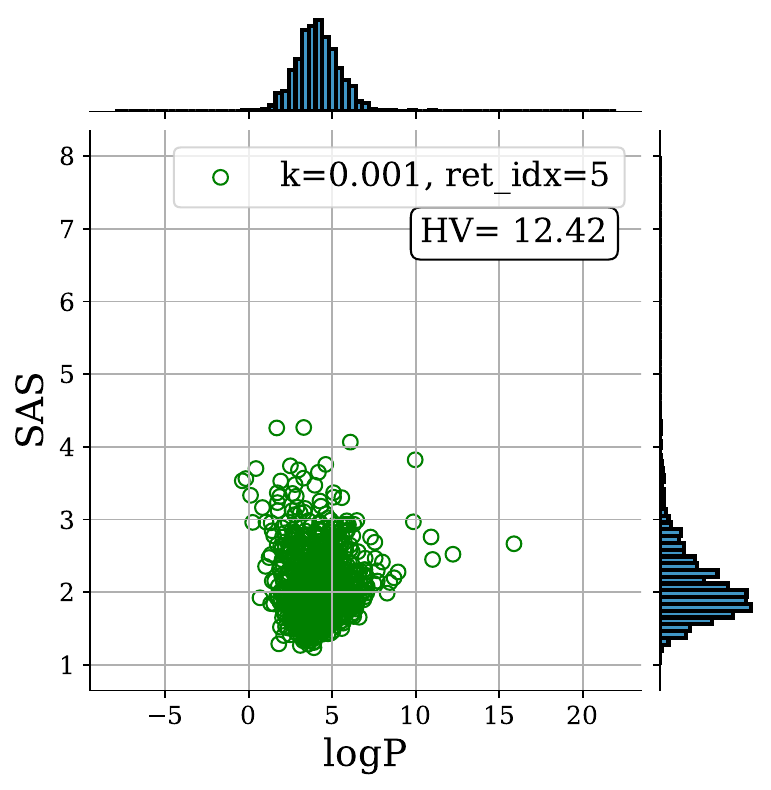}
    \end{minipage}%
    \begin{minipage}{.2\textwidth}
    \centering
    \includegraphics[width=\textwidth]{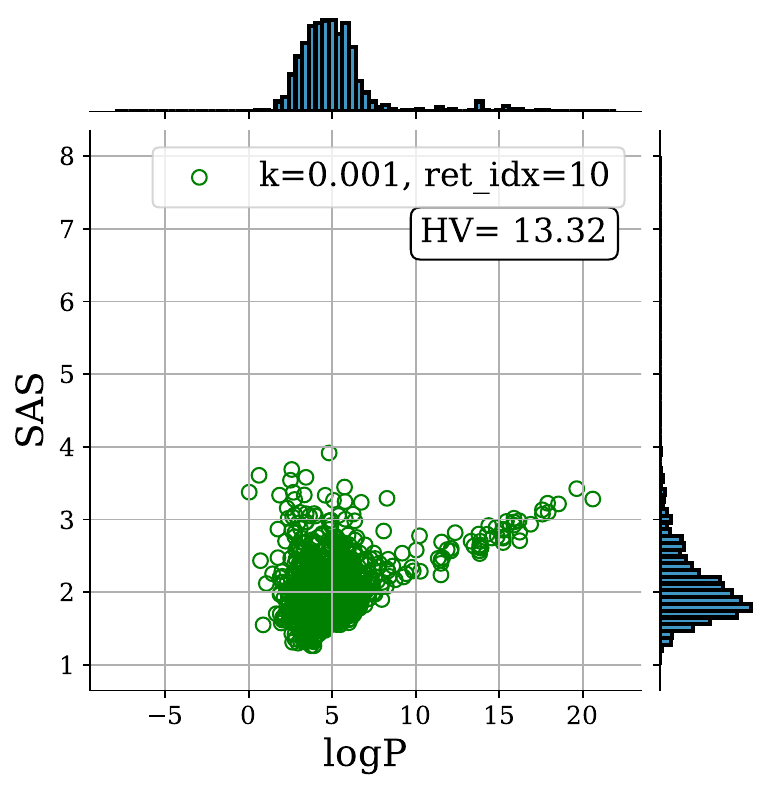}
    \end{minipage}
    
    \caption{\textbf{Effect of Bayesian optimization on the progression of the model in the objective space:} The first and second rows represent the scatter plots for random selection and Bayesian optimization strategy respectively. In each row, the objective space are for the training data, the molecules from the baseline model and retrained model after $1^{st}$, $5^{th}$ and $10^{th}$ weighted retraining for $k=10^{-3}$. Higher increase in hypervolume (HV) for Bayesian optimization strategy shows its effectiveness over random selection.}
    \label{scatter_t3}
\end{figure}

In previous sections, we generated novel molecules through random sampling in the latent space of the trained generative model, among which the top molecules were selected to augment the training data for the next cycle of weighted retraining. Although we adopted random sampling in order to demonstrate how the proposed multi-objective weighted retraining can enhance the sampling efficiency of the generative model for multiple target molecular properties, the overall efficacy of the GMD can be further improved by leveraging more sophisticated optimization techniques --  
such as Bayesian optimization (BO), genetic algorithms (GA), and particle swarm optimization (PSO)~\cite{PSO_cont_latent_space} 
-- to optimize the molecules in the latent space. 
To demonstrate this, we utilized BO to optimize molecules in the latent space of the generative model for logP and SAS. A single objective function was defined through scalarization, where logP (to be maximized) was penalized by SAS (to be minimized) after standardizing both values based on the mean and standard deviation of the respective values in the initial dataset. Based on the expected improvement acquisition function, we generate $50$ molecules and only unique samples are added to the training dataset.
Compared to the random generation strategy, the Bayesian optimization approach is computationally expensive since it requires training the GP surrogate model as well as optimizing the acquisition function. However, the BO approach is more sample efficient as it requires fewer property predictions to find the best data points to augment the training dataset. On the other hand, the random generation approach requires a larger number of evaluations (250 in our case) to select the best $50$ candidates for data augmentation.


Figure~\ref{scatter_t3} shows the results for $k=10^{-3}$. As shown in the top row, iterative weighted retraining with data augmentation through random sampling continues to shift the latent space distribution toward the desired direction, although not significantly even after the 10th iteration. On the other hand, data augmentation through Bayesian optimization shifts the latent space distribution much more effectively as can be seen in the bottom row of Figure~\ref{scatter_t3}. This is especially dramatic for logP, as after the 10th iterative retraining, the retrained generative model is capable of generating molecules with remarkably higher logP compared to those in the original training data. The candidate molecules suggested by the generative model tend to be somewhat biased towards the higher logP region with slightly higher SAS, which is likely an artifact due to the use of single-objective BO (SOBO) in this experiment, where the objective function was defined as a linear combination of logP and SAS. This may be addressed through different scalarization or the use of multi-objective BO (MOBO). Nonetheless, the example in Figure~\ref{scatter_t3} clearly shows the potential advantage of utilizing BO (or other advanced optimization schemes) for effective molecular optimization in the latent space, and thereby more effectively enhance the sampling efficiency of the generative model through the proposed multi-objective latent space optimization approach.

\section*{Discussion}


The presented framework for multi-objective optimization illustrates the potential of weighted retraining in bringing the molecule generation model in the expected multi-dimensional objective region. The JT-VAE model's latent space is optimized for different pairs of molecular properties and the effect of ranks are studied. We have found that the weight formulation from the ranks dictates a trade-off between the diversity and the shifts of the property distribution in the latent space. To verify the strength of our approach, similar experiments are repeated with having relatively poor training data. Even starting with no DRD2 active molecules, weighted retrained models still manage to configure its latent space for the active region. These outcomes are more pronounced given the fact that random selection from the latent space is used to propose the candidate molecules for retraining stages. To speed up the reshaping of the latent space, the Bayesian optimization is showed to be promising.

The ranking scheme of our framework is contingent on the robust property predictors. How well these surrogate models can tackle the unexplored chemical space is critical. Even if these models are less accurate, the retrained latent space can still be the exploration field for further screening. 
Given a reliable and fast approximate mapping from molecule to its property value, the weighted retraining approach can optimize the latent space jointly for more practical properties that are responsible for higher attrition rate of proposed drugs. With the availability of the surrogate models like protein-ligand binding score~\cite{binding_affinity}, inhibition of bile salt export pump~\cite{BSEP}, our approach can optimize the latent space in producing candidate drugs that are most likely to be active against specific target without causing possible damage to the patients. In terms of computational cost, retraining the generative network multiple times may be slightly expensive for larger network compared to the one we have used. However, with the application of distributed training~\cite{Rapid_covid}, the training time can be significantly mitigated.  Moreover, we have seen a decreasing trend in diversity of the molecules of the latent space with the increase in the shift of the property distribution. A diversity-oriented candidate selection strategy could be a better answer to this issue.

As shown in Fig.~\ref{scatter_t3}, the latent space optimization outcomes can be enhanced by adopting more effective sampling strategies in the latent space that can identify and suggest novel molecules with more desirable properties. In this study, this was demonstrated by replacing the random sampling scheme with Bayesian optimization. In fact, it may be possible to further improve the GMD through the proposed multi-objective LSO scheme by incorporating more sophisticated optimization techniques that can effectively explore the unknown landscape of the multiple objective functions under immense uncertainties. Optimal experimental design (OED)~\cite{Dehghannasiri2014tcbb, Dehghannasiri2015bmc,Hong2021,Woo2021} and active learning~\cite{Zhao2021NeurIPS,Zhao2021ICLR,Zhao2021AISTATS} techniques that build on objective-based uncertainty quantification (objective-UQ) based on MOCU (mean objective cost of uncertainty)~\cite{Yoon2013,Yoon2021} may provide practical solutions for such ``uncertainty-aware'' sampling in the latent space. These are topics of our ongoing investigation.

\section*{Methods}
\label{PF_rank}

\subsection*{Generative molecular design  using junction-tree variational autoencoder}
\label{wr_framework}
\vspace{10pt}

In this study, we have used the junction-tree variational autoencoder (JT-VAE) for investigating the multi-objective latent space optimization of deep generative models for GMD. While various deep generative models with latent space molecular representation have been proposed to date, JT-VAE is widely known for its high reconstruction accuracy when decoding latent samples into the original molecular space. Compared to other VAE models, where novel molecules sampled in the latent space often fail to decode into legitimate molecules, JT-VAE effectively addresses this issue by decomposing molecular graphs into a junction tree of chemical substructures. To suggest novel molecules, JT-VAE reconstructs the junction tree from the sampled latent embedding and assembles the chemical substructures into a molecular graph~\cite{JTVAE-paper}. The decoded tree structure serves as a scaffold guiding the generation of the molecular graph to reconstruct the molecule, resulting in a high fraction of valid molecules. However, we note that the proposed multi-objective latent space optimization scheme can be applied to various other types VAEs~\cite{Bombarelli_cont, kusner2017grammar} in a straightforward manner without any modification.

\subsection*{Non-dominated sorting and Pareto ranking}
\vspace{10pt}

In a multi-objective optimization problem, there may be no single solution that is optimal in terms of every objective. In practice, different objectives may conflict with each other, where optimizing one objective may result in a sub-optimal solution for one or more other objectives. The concept of Pareto optimality provides an effective way of addressing this issue and is widely utilized in the context of multi-objective optimization. Instead of finding the solution that optimizes a single objective, Pareto optimization aims to identify the Pareto optimal set, which is defined as the collection of all solutions that are not dominated by any other solution in the feasible solution space~\cite{deb_2004}. Consider the problem of jointly optimizing $K$ objective functions $f_1(\mathbf{x}), \cdots, f_K(\mathbf{x})$. Without loss of generality, we assume that the goal is to maximize all $K$ objective functions. Let $\mathbf{x}^i$ and  $\mathbf{x}^j$ be two points in the solution space. $\mathbf{x}^i$ is said to dominate $\mathbf{x}^j$ if $\mathbf{x}^i$ is as good as $\mathbf{x}^j$ in terms of all $K$ objectives (i.e., $f_k(\mathbf{x}^i) \geq f_k(\mathbf{x}^j)$, $\forall k=1,\cdots, K$) and if there is at least one objective such that $\mathbf{x}^i$ outperforms $\mathbf{x}^i$ (i.e., $\exists k$ s.t.  $f_k(\mathbf{x}^i) > f_k(\mathbf{x}^j)$).
The Pareto optimal points $\mathbf{x}$ in the solution space form the Pareto front (or Pareto  frontier) in the objective space. 

\begin{figure}[!ht]
    \centering
    \includegraphics[width=1\textwidth]{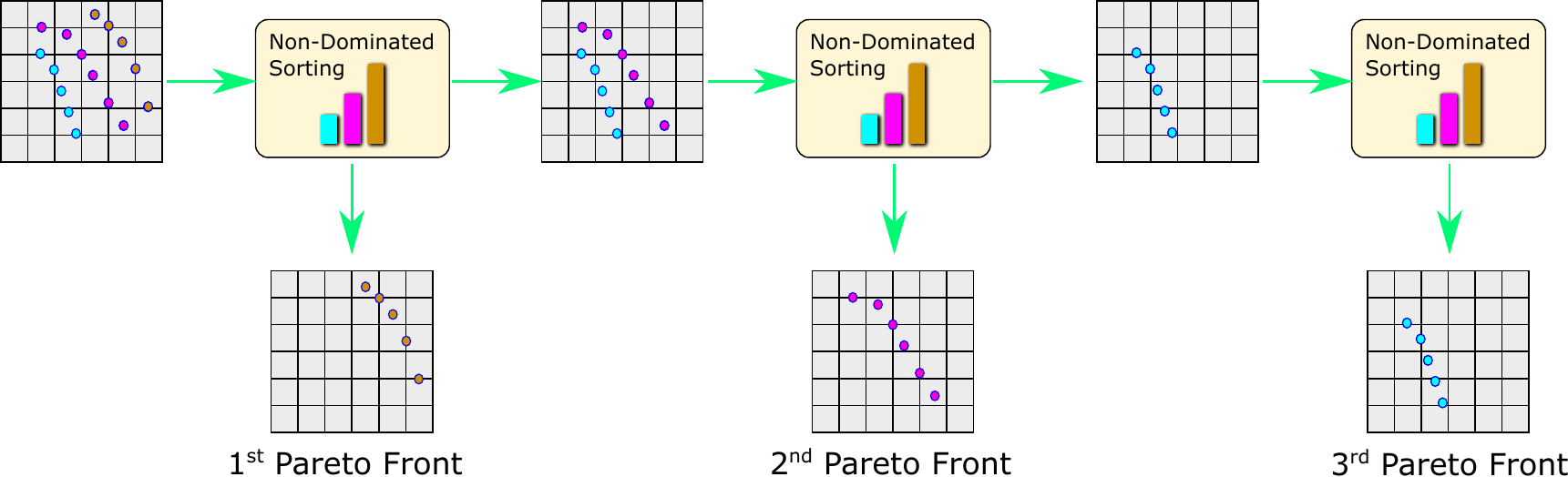}
    \caption{\textbf{Illustration of the Pareto front ranking process.} Suppose the main objective is to design novel molecules such that two target properties are jointly maximized. For weighted retraining of the generative model, we first identify the Pareto optimal molecules that are located on the Pareto front of the current dataset. All molecules in this first Pareto front are ranked 1. Next, these Pareto optimal molecules are removed from the dataset, after which we identify the second Pareto front among the remaining molecules. The Pareto optimal molecules in the second Pareto front are ranked 2. The process of removing the Pareto optimal molecules in the current dataset and the identification of the next Pareto front in the reduced dataset may be repeated until all molecules are ranked.
}
    \label{pf_rank_process}
\end{figure}

Considering all objectives simultaneously, all points in the Pareto front are equivalent to one another, as no point is either more preferable or less preferable than the others. As no point dominates any other point in the Pareto front, these points in the Pareto front may be assigned the same ranking.
The process of finding the Pareto front is summarized in \textbf{Algorithm~\ref{alg:1}}. Once the Pareto front is identified, we may remove these Pareto optimal points $\mathcal{P}_1$ from the dataset--i.e., ``peel off'' the first Pareto front--and move on to identify the next Pareto front $\mathcal{P}_2$ in the remaining dataset. This process may be repeated until all points in the dataset $\mathcal{D}$ are exhausted to find all possible Pareto optimal sets $\mathcal{P}_1, \mathcal{P}_2, \mathcal{P}_3, \cdots, \mathcal{P}_S$. Note that the sets $\mathcal{P}_1, \cdots, \mathcal{P}_S$ form a partition of $\mathcal{D}$ such that $\mathcal{P}_i \cap \mathcal{P}_j = \varnothing$ for $i\neq j$ and $\mathcal{D} = \cup_{i} \mathcal{P}_i$.
This Pareto ranking process is summarized in \textbf{Algorithm~\ref{alg:2}} 
 and illustrated in Figure~\ref{pf_rank_process}.
Based on these results, we may rank all data points such that all points in the $j$-th Pareto front are assigned the following ranking:
\begin{equation}
     rank_{\mathcal{D}}(\mathbf{x}) = \sum_{i=1}^{j-1} \vert \mathcal{P}_{i} \vert \qquad \forall \mathbf{x} \in \mathcal{P}_j
     \label{eq:rank_eqn}
\end{equation}
Despite practical differences from \eqref{eq:rank_eqn}, it is worth noting that similar multi-objective ranking schemes based on the concept of Pareto optimality have been previously explored in different contexts. For example, the work by Obayashi et~al.~\cite{obayashi1998niching} adopted a similar ranking scheme for a multi-objective genetic algorithm (MOGA) in order to identify solutions within the population of GA solutions that are nearly Pareto optimal.

\begin{algorithm}
\caption{Find the Pareto front by identifying the set of non-dominated data points}\label{alg:1}
\begin{algorithmic}
\Require N data points with their objective scores
\State Initialize $P' = \{1, 2, 3, ..., N\}$ \Comment{Set of non-dominated points, $P'$}
\State and $i \gets 1$
\While{$i \leq \vert P'\vert $}
\State Initialize $k \gets 0$ 
\For{(each $j\in P' \land j\neq i$)} 
\If{($x^{(j)}$ does not dominate $x^{(P'(i))}$ in any objective score)}
    \State $P' \gets P' \backslash \{j\} $
\ElsIf{$j < i$}
    \State $k \gets k+1$
\EndIf
\State $i \gets k + 1 $
\EndFor
\EndWhile

\end{algorithmic}
\end{algorithm}

\begin{algorithm}
\caption{Pareto front ranking of data points}\label{alg:2}
\begin{algorithmic}
\Require N data points with their objective scores
\State Initialize $P = \{1, 2, 3, ..., N\}$, $j=1$
\While{$\vert P \vert \neq 0$}
    \State \textbf{Step 1:} Find the non-dominated set, $P'$ from $P$ using \textbf{Algorithm~\ref{alg:1}}.
    \State \textbf{Step 2:} $P_j \gets P'$, $P \gets P\backslash P'$ and $j \gets j+1$ \Comment{$P_j$ : $j^{th}$ Pareto front}
\EndWhile

\end{algorithmic}
\end{algorithm}


\subsection*{Weighted retraining of generative models based on Pareto ranking}
\vspace{10pt}

We adopt the ranking scheme in~\eqref{eq:rank_eqn} for weighted retraining of the generative model (i.e., the JT-VAE in this study) to steer its latent space toward a region that is more sampling-efficient for enhanced molecules with multiple target properties. More specifically, we extend the weighted retraining scheme in~\cite{tripp2020sampleefficient} -- originally designed for latent space optimization based on a single objective -- to enable flexible multi-objective latent space optimization regardless of the number of objectives without any ad hoc scalarization of the objective function.
We calculate the weight for every data point $\mathbf{x}$ in the dataset $\mathcal{D}$ as follows
\begin{equation}
    w(\mathbf{x}, k , \mathcal{D}) = \frac{1}{kN + rank_{\mathcal{D}}(\mathbf{x})},
    \label{eq:weight}
\end{equation}
where this weight $w(\mathbf{x}, k , \mathcal{D})$ determines the influence of a given data point $\mathbf{x}$ on the training loss. As a consequence, a more desirable molecule with a higher Pareto ranking (i.e., a smaller $rank_{\mathcal{D}}(\mathbf{x})$) is assigned a larger weight $w(\mathbf{x}, k, \mathcal{D})$, thereby playing a more important role in retraining the generative model. $k$ is a hyper-parameter that adjusts the influence of the rank on the computed weight. A larger $k$ makes the weight distribution more uniform while a smaller $k$ assigns large weights to relatively fewer high-rank data points. $N = |\mathcal{D}|$ is the cardinality of the training set $\mathcal{D}$.

The iterative weighted retraining is performed as follows. We first start with the initial training dataset, $\mathcal{D}_{train} = \mathcal{D}_0$ and a null set $\mathcal{D}_{new} = \varnothing$. We retrain the model based on the given $\mathcal{D}_{train}$ after reweighting every data point (i.e., molecule) $\mathbf{x} \in \mathcal{D}_{train}$ by $w(\mathbf{x}, k , \mathcal{D}_{train})$ given by \eqref{eq:weight}. After the first weighted retraining with $\displaystyle \mathcal{D}_{train} = \mathcal{D}_0$, we randomly generate $250$ molecules from the latent space of this model, select the top $20\%$ among the generated molecules, which are then added to the set $\mathcal{D}_{new}$.
Next, we update the training dataset by augmenting it with the selected new molecules 
\begin{equation}
    \mathcal{D}_{train} \leftarrow \overline{\mathcal{D}}_0 \cup \mathcal{D}_{new},
\end{equation}
where $\overline{\mathcal{D}}_0$ may be either the initial training dataset $\mathcal{D}_0$ in its entirety, or a randomly sampled subset to reduce computation. In this study, we used $\overline{\mathcal{D}}_0$ by randomly  sampling 10\% of the molecules in the initial training data $\mathcal{D}_0$. In the next iteration, we can repeat the process by reweighting the molecules in $\mathcal{D}_{train}$, retraining the generative model, sampling new molecules in the latent space of the retrained model, and then selecting the top $20\%$ molecules and adding them to $\mathcal{D}_{new}$.
The weighted retraining cycle may be repeated a predetermined number of times or until a stopping criterion is reached. In this study, we repeated the cycle for 10 iterations to investigate the overall impact on enhancing the sampling efficiency of the generative model for suggesting novel molecules that simultaneously improve multiple target properties.


\subsection*{Molecular Property Predictors}

In this study, we considered four different molecular properties for validating the capability of the proposed multi-objective latent space optimization scheme and assessing its performance. The tested properties include: the partition coefficient (logP), synthesizability (SAS; synthetic accessibility score), natural product-likeness score (NP score) and inhibition of DRD2 (Dopamine receptor D2).

\paragraph{Partition coefficient (logP):}
As a quantitative measure of lipophilicity, the octanol-water partition coefficient (logP) is one of the standard properties for selecting potential drugs according to Lipinski's ``rule of 5"\cite{LIPINSKI19973}. We used the RDKit's rdkit.Chem.Crippen\cite{logP_paper} module to get the calculated logP values from SMILES representation.

\paragraph{Synthetic accessibility score (SAS):}
Synthetic accessibility score (SAS) of a molecule serves as a surrogate for quantifying the degree of difficulty in developing it. Although SAS does not account for the additional constraints the medicinal chemists may have in particular laboratory settings, e.g. restriction of using particular reagents, it is still useful in screening from large number of molecules for further evaluation. We obtained synthetic accessibility scores of molecules in our work from the RDKit-based implementation of SAS estimation method. In addition to the fragment score and the complexity penalty as in the original work\cite{SAS_paper}, this implementation includes the score based on the molecular symmetry.

\paragraph{Natural product-likeness score (NP score):}
Through the evolutionary selection process, the natural products often contain bio-active substructures which can be utilized in drugs\cite{HARVEY2008894}. The NP score for a molecule quantifies how much similarity its substructure has with the natural products. The higher score indicates that the molecular structure is more likely within the natural product space. We used the implementation from Ertl et al.\cite{NP_score_paper} which aggregates the individual scores for each fragments of a molecule.

\paragraph{Inhibition of DRD2 (Dopamine receptor D2):}
The dopamine receptor D2 has a long history of being used as the target protein for antipsychotic drugs\cite{drd2_structure}. More recent research findings\cite{drd2_cancer_1,drd2_cancer_2, drd2_cancer_3,drd2_cancer_4} demonstrate the effectiveness of DRD2 targeting drugs against a wide range of cancer cells. Thus searching for DRD2-inhibiting molecules provides us with an opportunity to showcase our proposed approach in a practical drug-discovery-like scenario.

We used an ML surrogate model in Olivecrona et al.~\cite{olivecrona2017} to predict the efficacy of a given molecule in inhibiting the activity of DRD2.
The surrogate model for predicting the activity against the dopamine type 2 receptor DRD2, is built as a binary Support Vector Machine (SVM) classifier with a radial basis function ($\displaystyle \gamma = 2^{-6}$). The Morgan Fingerprint, with radius 3 (FCFC6) computed by RDKit\cite{rdkit} are used as the input features for the SVM classifier. The probability of being active predicted by this model for each molecule is treated as a property to be maximized.
\begin{table}[!ht]
\centering
\begin{tabular}{@{}lcccc@{}}
\toprule
Dataset    & \multicolumn{1}{l}{Accuracy} & \multicolumn{1}{l}{AUC} & \multicolumn{1}{l}{Precision} & \multicolumn{1}{l}{Recall}\\ \midrule
Train      & 0.9998                & 0.9999  &0.9969 & 1.0      \\
Validation & 0.9807                 & 0.8745  &0.9747 & 0.7498        \\
Test       & 0.9842                 & 0.9074 &0.9770 & 0.8178             \\ \bottomrule
\end{tabular}
\caption{\label{tab:drd2_clf} \textbf{Performance of the DRD2 activity classifier used in this study.} The classifier predicts whether a given molecule may be an effective inhibitor of DRD2 (``active'') or not (``inactive'').} 
\end{table}\\
Since there is a class imbalance issue due to the smaller number of active molecules, Olivecrona et al.~\cite{olivecrona2017} split the active samples in such a way that the structural similarity among the samples from the train and test/validation dataset is less. We have used the same split dataset consisting of $7,218$ active and $100,000$ inactive molecules to train, test and validate the SVM model in Scikit-learn\cite{scikit-learn} (version 0.23.2) with a regularization parameter, $\displaystyle C = 2^7$. The overall performance of the DRD2 activity classifier used in this study is summarized in Table~\ref{tab:drd2_clf} for the training, validation, and test sets. Accuracy is defined as the fraction of samples correctly classified to be active or inactive. AUC denotes the area under the ROC (receiver operating characteristic) curve. Precision shows the ratio between correctly classified active molecules to all samples that are predicted to be active. Recall is the fraction of all active samples that are correctly classified.

\section*{Data and code availability}
The specific training and validation split of the ZINC dataset provided in \url{https://github.com/cambridge-mlg/weighted-retraining} is used in this work for retraining JT-VAE. The source code of the proposed multi-objective latent space optimization method can be downloaded from \url{https://github.com/nafizabeer/GMD-MO-LSO} and Zenodo\cite{abeer_2024_12730304}.

\section*{Acknowledgements}


This work represents a multi-institutional effort. Funding sources include the following: 
Federal funds from the National Cancer Institute, National Institutes of Health, and the Department of Health and Human Services, Leidos Biomedical Research Contract No. 75N91019D00024, Task Order 75N91019F00134 through the Accelerating Therapeutics for Opportunities in Medicine (ATOM) Consortium under CRADA TC02349. This work was supported in part by Brookhaven National Laboratory (BNL) LDRD No. 21-044.

Portions of this research were conducted with the advanced computing resources provided by Texas A\&M High Performance Research Computing (HPRC).

\section*{Author contributions statement}
FJA, BJY, NU initiated the project. BJY, NU proposed the idea. NA developed, refined, evaluated the method. MRW evaluated the DRD2 activity of the predicted molecules. NA, BJY analyzed the results and wrote the paper. All authors have edited and verified the paper.

\section*{Competing Interests}
Byung-Jun Yoon is a member of the Advisory Board of Patterns.
\section*{Materials and Correspondence}
This study did not generate any physical materials.\\
Byung-Jun Yoon is the lead contact for this study and can be reached at bjyoon@ece.tamu.edu

\bibliography{references, mocu}

\newpage
\setcounter{table}{0}
\renewcommand{\thetable}{S\arabic{table}}
\renewcommand\thesection{S\arabic{section}}
\doublespacing

\begin{center}
{\Large
\textbf{Supplementary information for ``Multi-Objective Latent Space Optimization of Generative Molecular Design Models''}
}
\\
A N M Nafiz Abeer$^{1}$, Nathan M. Urban$^{2}$, M Ryan Weil$^{3}$, Francis J. Alexander$^{4}$, 
and Byung-Jun Yoon$^{1,2,\ast}$
\\
\bf{1} Department of Electrical and Computer Engineering, Texas A\&M University, College Station, TX 77843, USA
\\
\bf{2} Computational Science Initiative, Brookhaven National Laboratory, Upton, NY 11973, USA
\\
\bf{3} Strategic and Data Science Initiatives, Frederick National Laboratory, Frederick, MD 21702, USA
\\
\bf{4} Computing, Environment and Life Sciences, Argonne National Laboratory, Lemont, IL 60439, USA
\\
$^{\ast}$ E-mail: bjyoon@ece.tamu.edu
\end{center}

\onehalfspacing


\section{Performance of weighted retraining for six pairs of properties}
\subsection{Effect of weight parameter $k$ with complete dataset}
\begin{table}[ht!]
\resizebox{\textwidth}{!}{%
\begin{tabular}{@{}llrrrrrrrr@{}}
\toprule
\multirow{2}{*}{Pair}           & \multirow{2}{*}{Property} & \multicolumn{1}{l}{\multirow{2}{*}{Training data}} & \multicolumn{1}{l}{\multirow{2}{*}{Initial Model}} & \multicolumn{6}{c}{Model after $10^{th}$ retraining with}                                                 \\ \cmidrule(l){5-10} 
                                & &\multicolumn{1}{l}{} & \multicolumn{1}{l}{}                               & \multicolumn{1}{l}{$k=10^{-1}$} & \multicolumn{1}{l}{$k=10^{-2}$} & \multicolumn{1}{l}{$k=10^{-3}$} & \multicolumn{1}{l}{$k=10^{-4}$} & \multicolumn{1}{l}{$k=10^{-5}$} & \multicolumn{1}{l}{$k=10^{-6}$} \\ \cmidrule(r){1-10}
\multirow{5}{*}{logP, SAS}      & logP                      & 2.4583        & 1.892                       & 2.9349 & 3.6783  & 4.0592  & 4.8422  & 5.2963 & 6.0017 \\ & logP (top 10\%)& 4.6321 & 4.2461 & 4.9848 & 5.6109& 5.9546 & 7.018 & 8.0196 & 8.9998\\
& SAS & 3.0535 & 3.2642 & 2.5012& 2.1818& 2.047    & 1.8387 & 1.4716 & 1.6566  \\
& SAS (top 10\%) & 1.9545 & 2.0544 & 1.5789 & 1.4669 & 1.4112 & 1.3475 & 1.004  & 1.1522\\
& Structural Diversity & \multicolumn{1}{l}{}  & 0.8356 & 0.7904 & 0.7512 & 0.7309 & 0.6773 & 0.6205  &0.6126\\ \cmidrule(lr){2-10}
\multirow{5}{*}{logP, NP score} & logP  & 2.4583 & 1.892 & 2.8972 & 3.2708 & 3.6407 & 4.0105  & 6.0847 & 6.3781 \\
& logP (top 10\%)           & 4.6321 & 4.2461   & 5.0467 & 5.5148 & 5.992 & 6.4334  & 8.1291 & 7.9213 \\
& NP\_score         & -1.3122 & -1.0196 & -0.7303 & -0.4854    & -0.2257 & 0.3444 & 1.5905 & 1.6276 \\
&NP\_score (top 10\%)&0.1482&0.1253&0.4336&0.8051&1.247&2.1975&2.3935&2.5946\\
 &Structural Diversity&\multicolumn{1}{l}{}&0.8356&0.8206&0.8231&0.8295&0.8347&0.7752&0.7529\\
                                \cmidrule(lr){2-10}
\multirow{5}{*}{NP score, SAS} 
&NP\_score&-1.3122&-1.0196&-0.7919&-0.5082&-0.2151&0.0759&0.9727&1.314\\
&NP\_score (top 10\%)&0.1482&0.1253&0.3403&0.86&1.3508&1.7946&2.7121&2.9791\\
&SAS&3.0535&3.2642&2.8367&2.6373&2.621&2.2284&2.3073&2.7489\\
&SAS (top 10\%)&1.9545&2.0544&1.6902&1.525&1.5213&1.3358&1.2841&1.4864\\
&Structural Diversity&\multicolumn{1}{l}{}&0.8356&0.8166&0.8122&0.8245&0.801&0.8042&0.8342
\\ \cmidrule(lr){2-10}
\multirow{5}{*}{logP, DRD2}     
&logP&2.4583&1.892&2.751&3.3405&3.7416&4.3834&8.3201&7.0397\\
&logP (top 10\%)&4.6321&4.2461&4.9568&5.5338&5.934&7.0267&10.7406&8.1715\\
&DRD2&0.0048&0.0066&0.0183&0.0374&0.0693&0.1477&0.5464&0.3587\\
&DRD2 (top 10\%)&0.0386&0.058&0.1572&0.3028&0.5151&0.8123&0.9971&0.9284\\
&Structural Diversity&\multicolumn{1}{l}{}&0.8356&0.8203&0.8133&0.7989&0.7812&0.6088&0.5845\\
\cmidrule(lr){2-10}
\multirow{5}{*}{DRD2, SAS}     
&DRD2&0.0048&0.0066&0.0085&0.021&0.0217&0.1063&0.424&0.5253\\
&DRD2 (top 10\%)&0.0386&0.058&0.0734&0.1937&0.1996&0.7922&1&0.9953\\
&SAS&3.0535&3.2642&2.7897&2.3898&2.2327&1.958&1.9466&2.1307\\
&SAS (top 10\%)&1.9545&2.0544&1.7315&1.4765&1.3778&1.2675&1.1678&1.2093\\
&Structural Diversity&\multicolumn{1}{l}{}&0.8356&0.8148&0.7932&0.776&0.7311&0.6509&0.4659\\
\cmidrule(lr){2-10}
\multirow{5}{*}{NP score, DRD2}
&NP\_score&-1.3122&-1.0196&-0.5209&-0.308&0.0339&0.7343&1.3774&1.8645\\
&NP\_score (top 10\%)&0.1482&0.1253&0.5555&0.8434&1.3966&2.3955&2.4835&2.9185\\
&DRD2&0.0048&0.0066&0.0179&0.0384&0.0412&0.0859&0.2032&0.0744\\
&DRD2 (top 10\%)&0.0386&0.058&0.1454&0.3052&0.3116&0.6029&0.933&0.5707\\
&Structural Diversity&\multicolumn{1}{l}{}&0.8356&0.8497&0.8518&0.8526&0.8531&0.8268&0.823\\
\bottomrule
\end{tabular}}
\caption{\textbf{Impact of weight parameter $k$ in proposed approach with complete dataset.} Average property values of the training data and $1000$ molecules randomly selected from the latent space of the initial model and the model learned after $10^{th}$ iteration of weighted retraining from starting with the complete dataset. For the top $10\%$ average value, molecules are ranked according to the corresponding property. The structural diversity is computed as the average structural distance based on ECFC4 fingerprints over all pairs in the set of $1000$ molecules.} 
\label{A_tab_1}
\label{A_tab_1}
\end{table}
\newpage
\subsection{Effect of weight parameter $k$ with reduced dataset (top $20\%$ molecules removed)}
\begin{table}[ht!]
\resizebox{\textwidth}{!}{%
\begin{tabular}{@{}llrrrrrrrr@{}}
\toprule
\multirow{2}{*}{Pair}           & \multirow{2}{*}{Property} & \multicolumn{1}{l}{\multirow{2}{*}{Training data}} & \multicolumn{1}{l}{\multirow{2}{*}{Initial Model}} & \multicolumn{6}{c}{Model after $10^{th}$ retraining with}                                                                     \\ \cmidrule(l){5-10} 
                                &                           & \multicolumn{1}{l}{}                               & \multicolumn{1}{l}{}                               & \multicolumn{1}{l}{$k=10^{-1}$} & \multicolumn{1}{l}{$k=10^{-2}$} & \multicolumn{1}{l}{$k=10^{-3}$} & \multicolumn{1}{l}{$k=10^{-4}$} & \multicolumn{1}{l}{$k=10^{-5}$} & \multicolumn{1}{l}{$k=10^{-6}$} \\ \cmidrule(r){1-10}
\multirow{5}{*}{logP, SAS} 
&logP&2.0875&1.6463&2.4422&2.7768&3.0227&3.2961&4.6374&5.7802\\
&logP (top 10\%)&3.8794&4.016&4.6657&4.8221&5.0974&5.4151&6.9437&8.3022\\
&SAS&3.2455&3.5983&3.1535&3.0309&2.8112&2.7935&2.077&1.5667\\
&SAS (top 10\%)&2.2257&2.2641&2.0878&1.9964&1.8365&1.8263&1.252&1.0853\\
&Structural Diversity&\multicolumn{1}{l}{}&0.8474&0.8319&0.822&0.82&0.8098&0.7441&0.5793\\ \cmidrule(l){2-10} 
\multirow{5}{*}{logP, NP score}&logP&2.2063&1.7481&2.305&2.4386&2.6125&2.8047&4.1676&7.7328\\
&logP (top 10\%)&4.0605&4.003&4.6038&4.7897&5.1549&5.3801&7.1442&10.6178\\
&NP\_score&-1.4805&-1.1313&-0.9178&-0.8948&-0.8349&-0.6123&-0.5973&0.0025\\
&NP\_score (top 10\%)&-0.4047&0.0479&0.1571&0.1556&0.2705&0.49&0.4912&1.1428\\
&Structural Diversity&\multicolumn{1}{l}{}&0.8404&0.8332&0.8312&0.8302&0.8376&0.7917&0.697\\ \cmidrule(l){2-10} 
\multirow{5}{*}{NP score, SAS}&NP\_score&-1.4945&-1.1036&-0.9777&-0.918&-0.8978&-0.8154&-0.4006&0.5817\\
&NP\_score (top 10\%)&-0.434&0.0183&0.1405&0.1754&0.2748&0.4079&1.2818&1.4811\\
&SAS&3.0918&3.4395&3.2631&3.164&3.0964&3.0436&2.3313&1.9048\\
&SAS (top 10\%)&2.1067&2.1432&2.0527&1.9997&1.9438&1.7133&1.294&1.1261\\ 
&Structural Diversity&\multicolumn{1}{l}{}&0.8411&0.8364&0.8338&0.8323&0.8375&0.8144&0.7737\\ \cmidrule(l){2-10} 
\multirow{5}{*}{logP, DRD2}&logP&2.2028&1.8422&2.2442&2.6281&2.8932&3.0949&3.7252&9.9092\\
&logP (top 10\%)&4.0734&4.1474&4.606&4.8051&5.196&5.3806&6.1771&21.8808\\
&DRD2&0.001&0.0028&0.0075&0.0159&0.0244&0.0216&0.0278&0.181\\
&DRD2 (top 10\%)&0.0053&0.0203&0.0607&0.1337&0.2061&0.1778&0.2206&0.8569\\
&Structural Diversity&\multicolumn{1}{l}{}&0.8351&0.8333&0.8319&0.831&0.8288&0.7987&0.7148\\ \cmidrule(l){2-10} 
\multirow{5}{*}{DRD2, SAS}&DRD2&0.0013&0.0044&0.0084&0.0084&0.0081&0.009&0.0356&0.1094\\
&DRD2 (top 10\%)&0.0071&0.0358&0.0732&0.072&0.07&0.0779&0.3287&0.7098\\
&SAS&3.1799&3.4301&3.313&3.1823&2.9907&2.9342&2.2998&1.6668\\
&SAS (top 10\%)&2.1636&2.1501&2.0841&2.0245&1.9232&1.8324&1.5063&1.1516\\
&Structural Diversity&\multicolumn{1}{l}{}&0.8452&0.8364&0.8334&0.8325&0.8315&0.8012&0.7252\\ \cmidrule(l){2-10} 
\multirow{5}{*}{NP score, DRD2}&NP\_score&-1.5318&-1.2088&-0.9943&-0.8575&-0.7821&-0.6855&-0.4658&0.7919\\
&NP\_score(top 10\%)&-0.5758&-0.1227&0.0749&0.2002&0.2949&0.414&0.8877&1.932\\
&DRD2&0.0008&0.0041&0.012&0.0116&0.0204&0.02&0.0562&0.2996\\
&DRD2 (top 10\%)&0.0041&0.035&0.1061&0.0966&0.1695&0.1692&0.4658&0.9653\\
&Structural Diversity&\multicolumn{1}{l}{}&0.8279&0.8367&0.837&0.8377&0.8416&0.8313&0.7327\\
\bottomrule

\end{tabular}}
\caption{\textbf{Impact of weight parameter $k$ in proposed approach with reduced dataset.} Average property values of the training data and $1000$ molecules randomly selected from the latent space of the initial model and the model learned after $10^{th}$ iteration of weighted retraining from starting with the reduced dataset (resulting from removal of top $20\%$ samples based on Pareto front rank). For the top $10\%$ average value, molecules are ranked according to the corresponding property. The structural diversity is computed as the average structural distance based on ECFC4 fingerprints over all pairs in the set of $1000$ molecules.}
\label{A_tab_2}
\end{table}

\newpage

\section{Comparison with scalarization-based optimization approach}
\label{scalarized_comp}
In this section, we compared our approach with the scalarization baseline of \cite{tripp2020sampleefficient} to showcase the improvement of our approach, i.e. the Pareto optimality based ranking of training datapoints.
For the scalarization baseline, the ranks of the training molecules are computed based on the weighted combination of their properties of interest where each property is standardized using the mean and standard deviation of the property values of complete training dataset. The weighted objective is defined as the summation of the standardized properties, where we take the negative of SAS property so that molecules with lower SAS gets higher weighted objective.

\cref{A_tab_3} shows the hypervolume measure corresponding to the Pareto front for the 500 molecules suggested over 10 iterations of weighted retraining for the six property pairs considered in this work. 
For each property, we performed weighted retraining based on the weighted objective as the scalarization baseline to empirically demonstrate the effectiveness of our approach where the rank of the molecules are governed by the Pareto optimality. For all property pairs, our approach outperforms the weighted objective method in most of the cases of $k$. In particular for lower values of $k$ (when rank is highly influenced by the property values), the improvement by our approach over the scalarization baseline is more pronounced.

\begin{table}[ht!]

\centering
\begin{tabular}{@{}ccrrrrr@{}}
\toprule
\multirow{2}{*}{\textbf{Property Pair}} &
  \multirow{2}{*}{\textbf{Rank Method}} &
  \multicolumn{5}{c}{Hypervolume} \\ \cmidrule(l){3-7} 
 &
   &
  \begin{tabular}[c]{@{}r@{}}$k=10^{-1}$\end{tabular} &
  $k=10^{-2}$ &
  $k=10^{-3}$ &
  $k=10^{-4}$ &
  $k=10^{-5}$ \\ \midrule
\multirow{2}{*}{logP, SAS} &
  weighted objective &
  5.5105 &
  \textbf{7.0641} &
  7.2546 &
  7.9114 &
  7.1866 \\
 &
  Ours &
  \textbf{6.0670} &
  6.8845 &
  \textbf{7.3066} &
  \textbf{8.0301} &
  \textbf{10.2210} \\ \midrule
\multirow{2}{*}{logP, NP score}                     & weighted objective & 6.7919 & 10.8326         & \textbf{17.2120} & 19.1626 & 19.2328 \\
 &
  Ours &
  \textbf{7.6259} &
  \textbf{11.7250} &
  15.5011 &
  \textbf{19.9497} &
  \textbf{22.9996} \\ \midrule
\multirow{2}{*}{NP score, SAS} &
  weighted objective &
  2.7127 &
  3.4984 &
  4.1356 &
  4.6484 &
  4.0521 \\
 &
  Ours &
  \textbf{2.9911} &
  \textbf{3.6366} &
  \textbf{4.2429} &
  \textbf{4.8607} &
  \textbf{5.3492} \\ \midrule
\multirow{2}{*}{logP, DRD2} &
  weighted objective &
  \textbf{2.6766} &
  2.9882 &
  3.5095 &
  3.7510 &
  2.9338 \\
 &
  Ours &
  2.2635 &
  \textbf{3.0061} &
  \textbf{3.8571} &
  \textbf{5.4800} &
  \textbf{7.3460} \\ \midrule
\multirow{2}{*}{DRD2, SAS} &
  weighted objective &
  0.8345 &
  1.0876 &
  1.2425 &
  1.3752 &
  1.3909 \\
 &
  Ours &
  \textbf{1.2639} &
  \textbf{1.3795} &
  \textbf{1.3993} &
  \textbf{1.5283} &
  \textbf{1.6890} \\ \midrule
\multicolumn{1}{l}{\multirow{2}{*}{NP score, DRD2}} & weighted objective & 1.3480 & \textbf{2.0743} & 2.0023           & 1.8920  & 2.0412  \\
\multicolumn{1}{l}{} &
  Ours &
  \textbf{1.6305} &
  1.9528 &
  \textbf{2.1722} &
  \textbf{3.5243} &
  \textbf{3.2791} \\ \bottomrule
\end{tabular}
\caption{ \textbf{Hypervolume of property space dominated by the Pareto front for the $500$ molecules suggested over 10 weighted retraining iterations for six property pairs.} For the ``weighted objective" rank method, the ranking of the molecules at each retraining is based on a scalar quantity which is the summation of standardized properties using the mean and standard deviation for the corresponding properties in the training dataset. On the other hand, our approach utilizes the Pareto front rank based on both properties.  The reference point for hypervolume computation is set to the average property value of the complete training dataset.}
\label{A_tab_3}
\end{table}

\newpage
\section{Optimization for three properties}
\label{three_prop}
To demonstrate the effectiveness of our approach more than two properties, we considered the latent space optimization for logP, SAS and DRD2 where all properties except SAS are expected to be maximized. We followed the same procedure as the bi-objective cases, and ran the weighted retraining for $k=10^{-3}, 10^{-4}, 10^{-5}$. \cref{A_tab_4} the hypervolume of the properties for 500 molecules suggested over 10 weighted retraining iterations using two methods for determining the rank of training molecules: one is based on weighted objective \cite{tripp2020sampleefficient} and the other one is our approach which uses the Pareto front rank derived from all three properties. The higher hypervolume in all three values of $k$ indicates the efficacy of our approach over the weighted objective-based ranking.

\begin{table}[ht!]
\centering

\begin{tabular}{@{}ccrrr@{}}
\toprule
\multirow{2}{*}{\textbf{Property Triplet}} & \multirow{2}{*}{\textbf{Method}} & \multicolumn{1}{l}{} & \multicolumn{1}{l}{Hypervolume} & \multicolumn{1}{l}{} \\ \cmidrule(l){3-5} 
                                 &                    & $k=10^{-3}$           & $k=10^{-4}$         & $k=10^{-5}$        \\ \midrule
\multirow{2}{*}{logP, SAS, DRD2} & weighted objective & 2.8937          & 3.9735          & 3.6107          \\
                                 & Ours               & \textbf{3.7959} & \textbf{4.3234} & \textbf{5.0013} \\ \bottomrule
\end{tabular}
\caption{ \textbf{Hypervolume of property space dominated by the Pareto front for the $500$ molecules suggested over 10 weighted retraining iterations for logP, SAS and DRD2.} For the ``weighted objective" rank method, the ranking of the molecules at each retraining is based on a scalar quantity which is the summation of properties standardized using the mean and standard deviation for the corresponding properties in the training dataset. On the other hand, our approach utilizes the Pareto front rank based on all three properties.  The reference point for hypervolume computation is set to the average property value of the molecules in complete training dataset.}
\label{A_tab_4}
\end{table}

\section{Impact of retraining on molecule reconstruction accuracy}

Our goal is to make the generative model, JT-VAE more biased to produce molecules with better properties and we achieve this goal by optimizing the JT-VAE parameters in such a way that the model primarily learns to reconstruct molecules that have better properties than others in the training dataset. 

While the weighted retraining makes the latent space more biased toward the molecules with desired properties, it also means that the retrained model has poor reconstruction performance for molecules with low-scoring properties.
We empirically investigated this effect by comparing the reconstruction performance of the pre-trained JT-VAE model as well as the retrained models corresponding to $k\in \{10^{-1}, 10^{-2}, 10^{-3}, 10^{-4}, 10^{-5}\}$ for the property pair: logP and SAS.
Specifically, we compute the negative log-likelihood (nLL) over the validation dataset for pre-trained JT-VAE and the retrained models corresponding to 5 cases of $k$. The model with lower nLL has better performance in constructing the molecules of the validation dataset.
For each model, we repeat the nLL computation 5 times, and their average and standard deviation are reported in \cref{A_tab_5}. 

\begin{table}[ht]
\centering 
\begin{tabular}{@{}cc@{}}
\toprule
\textbf{JT-VAE Model} & \textbf{nLL (5 repetitions)} \\ \midrule
Pre-trained           & 1.4816 (0.0040)               \\
$k=10^{-1}$             & 3.4983 (0.0037)              \\
$k=10^{-2}$             & 4.6442 (0.0033)              \\
$k=10^{-3}  $           & 6.1943 (0.0051)              \\
$k=10^{-4} $            &    11.0966 (0.0031)                          \\
$k=10^{-5}$             &      20.1343 (0.0051)                        \\ \bottomrule
\end{tabular}
\caption{ \textbf{Reconstruction performance on the validation dataset.} Negative log-likelihood (nLL) for pre-trained JT-VAE and the retrained models using our proposed approach for $k\in \{10^{-1}, 10^{-2}, 10^{-3}, 10^{-4}, 10^{-5}\}$ for property pair: logP and SAS. The metric is computed over the same validation split 5 times, and their average and standard deviation are reported.}
\label{A_tab_5}
\end{table}
The pre-trained model shows the best reconstruction performance (lowest nLL) since it was trained to reconstruct wide range of molecules, unlike the retrained models. 
As we retrain the generative model with the smaller value of $k$ which means more weights on the molecules with better properties, the retrained model becomes biased towards high-scoring molecules. Since validation split contains molecules with a broad range of properties, for the retrained models with smaller $k$, we are seeing higher nLL which corresponds to the poor reconstruction accuracy for the molecules within the validations dataset.

\section{Comparison with MARS \cite{mars}}

In this section, we compare our approach with Xie et al. \cite{mars}'s proposed sampling-based technique -- MARS. Since this approach trains the molecule generator unit, i.e. molecular graph editing model in an online fashion, it has a certain advantage over the deep generative models (such as the VAE-based models that we have focused on in the study) which are trained in self-supervised fashion without considering the properties of the training molecules. We have used the implementation of \cite{mars} to generate $500$ molecules for the same combination of properties considered in our work. 
For the scoring function of logP and NP score, the property value is standardized using the same mean and standard deviation that we used for weighted objective in \cref{scalarized_comp,three_prop}. We also have followed their implementation to normalize the SAS into $[0,1]$ interval where the higher score corresponds to the lower SAS.

\begin{table}[!ht]
\centering 
\begin{tabular}{@{}cccc@{}}
\toprule
\multirow{2}{*}{Properties} & \multicolumn{3}{c}{Hypervolume} \\ \cmidrule(l){2-4} 
                            & MARS \cite{mars}           & $k=10^{-5}$    & $k=10^{-4}$    \\ \midrule
logP, SAS                   & \textbf{12.6780}        & 10.2210    & 8.0301    \\
logP, NP score              & \textbf{45.0480}        & 22.9996  & 19.9497      \\
NP score, SAS               & 3.3847         & \textbf{5.3492}    & 4.8607     \\
logP, DRD2                  & 6.6405         & \textbf{7.3460}   & 5.4800      \\
DRD2, SAS                   & 1.5571         & \textbf{1.6890}   & 1.5283      \\
NP score, DRD2              & 3.3672         & 3.2791      & \textbf{3.5243}   \\
logP, SAS, DRD2             & \textbf{5.1056}         & 5.0013      & 4.3234   \\ \bottomrule
\end{tabular}
\caption{ \textbf{Comparison between MARS and our approach for different combinations of properties.} Hypervolume of property space dominated by the Pareto front for the $500$ molecules suggested using MARS \cite{mars} and our proposed approach of 10 weighted retraining iterations with $k\in \{10^{-5}, 10^{-4}\}$ for different combination of properties.
The reference point for hypervolume computation is set to the average property value of the molecules in complete training dataset.} 
\label{mars_exp}
\end{table}

\cref{mars_exp} shows the hypervolume metric for $500$ molecules suggested by MARS and our Pareto front rank approach with $k\in \{10^{-5}, 10^{-4}\}$. 
For property pairs -- (logP, DRD2), (DRD2, SAS), (NP score, SAS) and (NP score, DRD2) our proposed approach produced more Pareto-optimal molecules than MARS. On the other hand, 
the MARS approach tends to show superior performance over ours when logP is one of the properties. As mentioned in \cite{mars}, it is easier for MARS to optimize for logP by generating larger molecules whereas the generative models like JT-VAE in our work struggle to do so.
However, our proposed approach enables the JT-VAE model to outperform MARS for the property pair logP and DRD2.
We speculate that, for challenging objectives like DRD2, MARS loses its advantage in logP because simply adding more molecular units does not translate to increased DRD2 inhibition probability.
This is also reflected in the case of the property triplet where MARS and our approach perform similarly.

\end{document}